\documentclass{article}
\usepackage{ijcai20}

\usepackage{times}
\usepackage{soul}
\usepackage{url}
\usepackage{hyperref}
\usepackage[utf8]{inputenc}
\usepackage[small]{caption}
\usepackage{graphicx}
\usepackage{amsmath}
\usepackage{amsthm}
\usepackage{booktabs}
\usepackage{algorithm}
\usepackage{algorithmic}
\usepackage[capitalise]{cleveref}
\usepackage{amssymb}
\usepackage{centernot}
\usepackage{longtable}
\usepackage{caption}
\usepackage{subcaption}
\usepackage{enumitem}
\usepackage{xcolor}
\usepackage{bibunits}

\defaultbibliography{paper}
\defaultbibliographystyle{abbrvnamed}

\newcommand{\R}{\mathbb{R}}

\renewcommand{\vec}[1]{\mathbf #1}
\newcommand{\x}{\mathbf{x}}
\newcommand{\w}{\mathbf{w}}
\newcommand{\X}{\mathcal{X}}

\newcommand{\lm}{\boldsymbol{\lambda}}
\newcommand{\tht}{\boldsymbol{\theta}}
\newcommand{\ph}{\boldsymbol{\phi}}

\newtheorem{proposition}{Proposition}

\title{Mixed-Variable Bayesian Optimization}

\author{
Erik Daxberger\footnote{Equal contribution. $^\dagger$Work done while at ETH Zurich. %
}$^{,\dagger,1,2}$
\and
Anastasia Makarova$^{*,3}$\and
Matteo Turchetta$^{2,3}$\And
Andreas Krause$^3$
\affiliations
$^1$Department of Engineering, University of Cambridge\\
$^2$Max Planck Institute for Intelligent Systems, T\"{u}bingen\\
$^3$Department of Computer Science, ETH Zurich\\
\emails
ead54@cam.ac.uk,\{anmakaro,matteo.turchetta,krausea\}@inf.ethz.ch
}

\begin{document}

\maketitle

\begin{bibunit}
\begin{abstract}
    The optimization of expensive to evaluate, black-box, mixed-variable functions, i.e. functions that have continuous and discrete inputs, is a difficult and yet pervasive problem in science and engineering.
    In Bayesian optimization (BO), special cases of this problem that consider fully continuous or fully discrete domains have been widely studied. However, few methods exist for mixed-variable domains and none of them can handle discrete constraints that arise in many real-world applications.
    In this paper, we introduce \textsc{MiVaBo}, a novel BO algorithm for the efficient optimization of mixed-variable functions combining a linear surrogate model based on expressive feature representations with Thompson sampling. We propose an effective method to optimize its acquisition function, a challenging problem for mixed-variable domains, making \textsc{MiVaBo} the first BO method  that can handle complex constraints over the discrete variables. Moreover, we provide the first convergence analysis of a mixed-variable BO algorithm. Finally, we show that \textsc{MiVaBo} is significantly more sample efficient than state-of-the-art mixed-variable BO algorithms on several hyperparameter tuning tasks, including the tuning of deep generative models.

\end{abstract}

\section{Introduction}
\label{sec:intro}
Bayesian optimization (BO) \cite{movckus1975} is a well-established paradigm to optimize costly-to-evaluate, complex, black-box objectives that has been successfully applied to many scientific domains. Most of the existing BO literature focuses on objectives that have purely \emph{continuous} domains, such as those arising in tuning of continuous hyperparameters of machine learning algorithms, recommender systems, and preference learning \cite{shahriari16}. More recently, problems with purely \emph{discrete} domains, such as food safety control and model-sparsification in multi-component systems \cite{baptista2018} have been considered.

However, many real-world optimization problems in science and engineering are of \emph{mixed-variable} nature, involving \emph{both} continuous and discrete input variables, and exhibit complex constraints. For example, tuning the hyperparameters of a convolutional neural network involves both continuous variables, e.g., learning rate and momentum, and discrete ones, e.g., kernel size, stride and padding. Also, these hyperparameters impose validity constraints, as some combinations of kernel size, stride and padding define invalid networks. Further examples of mixed-variable, potentially constrained, optimization problems include sensor placement \cite{krause2008}, drug discovery \cite{negoescu2011}, optimizer configuration \cite{hutter2011} and many others. Nonetheless, only few BO methods can address the unconstrained version of such problem and no existing method can handle the constrained one. This work introduces the first algorithm that can efficiently optimize mixed-variable functions subject to known constraints with provable convergence guarantees.
\paragraph{Related Work.} 
Extending \emph{continuous} BO methods
\cite{shahriari16}
to mixed inputs requires ad-hoc relaxation methods to map the problem to a fully continuous one and rounding methods to map the solution back. This ignores the original domain structure, makes the solution quality dependent on the relaxation and rounding methods, and makes it hard to handle discrete constraints. 
Extending \emph{discrete} BO methods \cite{baptista2018,oh2019} to mixed inputs requires a discretization of the continuous domain part, the granularity of which is crucial: If it is too small, the domain becomes prohibitively large; if it is too large, the domain may only contain poorly performing values of the continuous inputs.
Few BO methods address the mixed-variable setting.
SMAC \cite{hutter2011} uses a random forest surrogate model. However, its frequentist uncertainty estimates may be too inaccurate to steer the sampling.
TPE \cite{bergstra2011a} uses
kernel density estimation to find inputs that will likely improve upon and unlikely perform worse than the incumbent solution.
While SMAC and TPE can handle hierarchical constraints, they cannot handle more general constraints over the discrete variables, e.g., cardinality constraints. They also lack convergence guarantees.
Hyperband (HB) \cite{li2016} uses cheap but less accurate approximations of the objective to dynamically allocate resources for function evaluations. BOHB \cite{falkner2018} is the model-based counterpart of HB, based on TPE. They thus extend existing mixed-variable methods to the multi-fidelity setting rather than proposing new ones, which is complementary to our approach, rather than in competition with it.
\cite{garrido2018} propose a Gaussian process kernel to model discrete inputs without rounding bias.
Their method lacks guarantees and cannot handle discrete constraints. We instead use discrete optimizers for the acquisition function, which avoid bias by only making integer evaluations.
Finally, while \cite{hernandez2015c,gardner2014,sui2015safe} extend continuous BO methods to handle unknown constraints, no method can handle known discrete constraints in a mixed-variable domain.

\paragraph{Contributions.} We introduce \textsc{MiVaBo}, the first BO algorithm for efficiently optimizing mixed-variable functions subject to known linear and quadratic integer constraints, encompassing many of the constraints present in real-world domains (e.g. cardinality, budget and hierarchical constraints).
It relies on a linear surrogate model that decouples the continuous, discrete and mixed components of the function using an expressive feature expansion (Sec.~\ref{subsec:model}). We exploit the ability of this model to efficiently draw samples from the posterior over the objective (Sec.~\ref{subsec:inference}) by combining it with Thompson sampling, and show how to optimize the resulting constrained acquisition function (Sec.~\ref{subsec:acquisition}).
While in continuous BO, optimizing the acquisition function is difficult but has well-established solutions, this is not true for mixed-variable spaces and doing this efficiently and accurately is a key challenge that hugely impacts the algorithm's performance. We also provide the first convergence analysis of a mixed-variable BO algorithm (Sec.~\ref{subsec:regret_bounds}). Finally, we demonstrate the effectiveness of \textsc{MiVaBo} on a set of complex hyperparameter tuning tasks, where it outperforms state-of-the-art methods and is competitive with human experts (Sec.~\ref{sec:experiments}).

\section{Problem Statement}
\label{sec:problem}
We consider the problem of optimizing an unknown, costly-to-evaluate function defined over a mixed-variable domain, accessible through noisy evaluations and subject to known linear and quadratic constraints. Formally, we aim to solve 
\vspace{-3mm}
\begin{equation} 
    \label{eq:ProblemDefinition} 
	{\min}_{\x \in \X} ~f(\x) \quad \text{s.t. } ~g^c(\x) \geq 0, ~g^d(\x) \geq 0, 
\end{equation}
where $\X \subseteq \X^c \times \X^d$ with continuous subspace $\X^c$ and discrete subspace $\X^d$.
Both constraints $g^c(\x) \geq 0$ over $\X^c$ and $g^d(\x) \geq 0$ over $\X^d$ are known, and specifically $g^d(\x)$ are linear or quadratic.
We assume, that the domain of the continuous inputs is box-constrained and can thus, w.l.o.g., be scaled to the unit hypercube, $\X^c = [0, 1]^{D_c}$.
We further assume, w.l.o.g., that the discrete inputs are binary, i.e., vectors $\x^d \in \X^d = \{0,1\}^{D_d}$ are vertices of the unit hypercube.
This representation can effectively capture the domain of any discrete function.
For example, a vector $\x^d = [x^d_i]_{i=1}^{D_d} \in \X^d$ can encode a subset $A$ of a ground set of $D_d$ elements, such that $x^d_i = 1 \Leftrightarrow a_i \in A$ and $x^d_i = 0 \Leftrightarrow a_i \notin A$, yielding a set function.
Alternatively, $\x^d \in \X^d$ can be a binary encoding of integer variables, yielding a function defined over integers.
\paragraph{Background.} BO algorithms are iterative black-box optimization methods which, at every step $t$, select an input $\x_t \in \X$ and observe a noise-perturbed output $y_t \triangleq f(\x_t) + \epsilon$ with $\epsilon \overset{\text{iid}}{\sim} \mathcal{N}(0, \beta ^{-1})$,  $\beta > 0$.
As evaluating $f$ is costly, the goal is to query inputs based on past observations to find a global minimizer
$\x_* \in \operatorname{arg\, min}_{\x \in \X} f(\x)$
as efficiently and accurately as possible.
To this end, BO algorithms leverage two components: (i) a \emph{probabilistic function model} (or \emph{surrogate}), that encodes the belief about $f$ based on the observations available, and (ii) an \emph{acquisition function} $\alpha:\mathcal{X}\rightarrow \mathbb{R}$ that expresses the informativeness of input $\x$ about the location of $\x_*$, given the surrogate of $f$. 
Based on the model of $f$, we query the best input measured by the acquisition function, then update the model with the observation and repeat this procedure.
The goal of the acquisition function is to simultaneously learn about inputs that are likely to be optimal and about poorly explored regions of the input space, i.e., to trade-off exploitation against exploration.
Thus, BO reduces the original hard black-box optimization problem to a series of cheaper problems $\x_t \in \operatorname{arg\, max}_{\x \in \X} \alpha_t(\x)$. However, in our case, these optimization problems involve mixed variables and exhibit linear and quadratic constraints and are thus still challenging.
We now present \textsc{MiVaBo}, an algorithm to efficiently solve the optimization problem in \cref{eq:ProblemDefinition}.  

\section{\textsc{MiVaBo} Algorithm}
\label{sec:main}
We first introduce the linear model used to represent the objective 
(Sec.~\ref{subsec:model})
and describe how to do inference with it
(Sec.~\ref{subsec:inference}).
We then show how to use Thompson sampling
to query informative inputs
(Sec.~\ref{subsec:acquisition})
and, finally, provide a bound on the regret incurred by \textsc{MiVaBo}.
(Sec.~\ref{subsec:regret_bounds}).

\subsection{Model}
\label{subsec:model}
We propose a surrogate model that accounts for both discrete and continuous variables in a principled way, while balancing two conflicting goals: Model expressiveness versus feasibility of Bayesian inference and of the constrained optimization of the mixed-variable acquisition function. 
Linear models defined over non-linear feature mappings, $f(\x)=\w^\top \ph(\x)$, are a class of flexible parametric models that strike a good trade-off between model capacity, interpretability and ease of use through the definition of features $\ph:\X\rightarrow \R^M$. 
While the complexity of the model is controlled by the number of features, $M$, its capacity depends on their definition. Therefore, to make the design of a set of expressive features more intuitive, we treat separately the contribution to the objective $f$ from the discrete part of the domain, from the continuous part of the domain, and from the interaction of the two,
\begin{equation}
\label{eq:f}
    f(\x) = \textstyle\sum_{j\in \{d,c,m\}}{\w^j}^\top \ph^j(\x^j)
\end{equation}
where, for $j\in\{d, c, m\}$, $\ph^j(\x^j) = [\phi^j_i(\x^j)]_{i=1}^{M_j} \in \mathbb{R}^{M_j}$ and $\w^j \in \mathbb{R}^{M_j}$ are the feature and weight vector for the $d$iscrete, $c$ontinuous and $m$ixed function component, respectively.

In many real-world domains, a large set of features can be discarded \emph{a priori} to simplify the design space. It is common practice in high-dimensional BO to assume that only low-order interactions between the variables contribute significantly to the objective, which was shown for many practical problems \cite{rolland2018,mutny2018}, including deep neural network hyperparameter tuning \cite{hazan2017}. Similarly, we focus on features defined over small subsets of the inputs.
Formally, we consider $\ph(\x) = [\phi_{k}(\x_{k})]_{k=1}^{M}$, where $\x_{k}$ is a subvector of $\x$ containing exclusively continuous or discrete variables or a mix of both. Thus, the objective $f(\x)$ can be decomposed into a sum of low-dimensional functions $f_{k}(\x_{k}) \triangleq w_k \phi_k(\x_{k})$ defined over subspaces $\X_{k} \subseteq \X$ with
$\text{dim}(\X_{k}) \ll \text{dim}(\X)$.
This defines a \emph{generalized additive model} \cite{rolland2018,hastie2017}, where the same variable can be included in multiple subvectors/features. The complexity of this model is controlled by the \emph{effective dimensionality} (ED) of the subspaces, which is crucial under limited computational resources.
In particular, let $\bar{D}_d \triangleq \max_{k \in [M]} \text{dim}(\X^d_{k})$ denote the ED of the discrete component in \cref{eq:f}, i.e. the dimensionality of the largest subspace that exclusively contains discrete variables. Analogously, $\bar{D}_c$ and $\bar{D}_m$ denote the EDs of the continuous and mixed component, respectively.
Intuitively, the ED corresponds to the maximum order of the variable interactions present in $f$.
Then, the number of features $M \in \mathcal{O} \big(D_d^{\bar{D}_d} + D_c^{\bar{D}_c} + (D_d+D_c)^{\bar{D}_m} \big)$ scales exponentially in the EDs only (as modeling up to $L$-th order interactions of $N$ inputs requires $\sum_{l=0}^L \binom{N}{l} \in \mathcal{O}(N^L)$ terms), which are usually small, even if the true dimensionality is large.
\paragraph{Discrete Features $\ph^d$.} We aim to define features $\phi^d$ that can effectively represent the discrete component of \cref{eq:f} as a linear function, which should generally be able to capture arbitrary interactions between the discrete variables.
To this end, we consider all subsets $S$ of the discrete variables in $\X^d$ (or, equivalently, all elements $S$ of the powerset $2^{\X_d}$ of $\X_d$) and define a monomial $\prod_{j \in S} x^d_j$ for each subset $S$ (where for $S = \emptyset$, $\prod_{j \in \emptyset} x^d_j = 1$).
We then form a weighted sum of all monomials to yield the multi-linear polynomial
${\w^d}^\top \ph^d(\x^d) = \sum_{S \in 2^{\X_d}} w_S \prod_{j \in S} x^d_j$.
This functional representation corresponds to the Fourier expansion of a \emph{pseudo-Boolean function} (PBF) \cite{boros2002}.
In practice, an exponential number of features can be prohibitively expensive and may lead to high-variance estimators as in BO one typically does not have access to enough data to robustly fit a large model. 
Alternatively, \cite{baptista2018,hazan2017} empirically found that a second-order polynomial in the Fourier basis provides a practical balance between expressiveness and efficiency, even when the true function is of higher order. In our model, we also consider quadratic PBFs,
${\w^d}^\top \ph^d(\x^d) = w_\emptyset + \sum_{i=1}^n w_{\{i\}} x^d_i + \sum_{1 \leq i < j \leq n} w_{\{i,j\}} x^d_i x^d_j$,
which induces the discrete feature representation
$\ph^d(\x^d) \triangleq [ 1, \{x^d_i\}_{i=1}^{D_d}, \{x^d_i x^d_j\}_{1 \leq i < j \leq D_d} ]^\top$ and
reduces the number of model weights to $M_d \in \mathcal{O}(D_d^2)$.
\paragraph{Continuous Features $\ph^c$.} In BO over continuous spaces, most approaches are based on 
Gaussian process (GP) models \cite{williams2006gaussian} due to their flexibility and ability to capture large classes of continuous functions.
To fit our linear model formulation, we leverage GPs' expressiveness by modeling the continuous part of our model in \cref{eq:f} using feature expansions $\ph^c(\x^c)$ that result in a finite linear approximation of a GP.
One simple, yet theoretically sound, choice
is the class of Random Fourier Features (RFFs) \cite{rahimi2008}, which use Monte Carlo integration for a randomized approximation of a GP.
Alternatively, one can use Quadrature Fourier Features \cite{mutny2018}, which instead use numerical integration for a deterministic approximation, which is particularly effective for problems with low effective dimensionality.
Both feature classes were successfully used in BO \cite{jenatton2017,mutny2018}.
In our experiments, we use RFFs approximating a GP with a squared exponential kernel, which we found to best trade off complexity vs.\ accuracy in practice.
\paragraph{Mixed Features $\ph^m$.} The mixed term should capture as rich and realistic interactions between the discrete and continuous variables as possible, while keeping model inference and acquisition function optimization efficient.  
To this end, we stack products of all pairwise combinations of features of the two variable types, i.e. $\ph^m(\x^d, \x^c) \triangleq [\phi^d_i(\x^d) \cdot \phi^c_j(\x^c)]^\top_{1 \leq i \leq M_d, 1 \leq j \leq M_c}$.
This formulation provides a good trade-off between modeling accuracy and computational complexity.
In particular, it allows us to reduce $\ph^m$ to the discrete feature representation $\ph^d$ when conditioned on a fixed assignment of continuous variables $\ph^c$ (and vice versa).
This property is crucial for optimizing the acquisition function, as it allows us to optimize the mixed term of our model by leveraging the tools for optimizing the discrete and continuous parts individually.
The proposed representation contains $M_d M_c$ features, resulting in a total of $M = M_d + M_c + M_d M_c$.
To reduce model complexity, prior knowledge about the problem can be incorporated into the construction of the mixed features. In particular, one may consider the following approaches.
Firstly, one can exploit a known interaction structure between variables, e.g., in form of a dependency graph, and ignore the features that are known to be irrelevant. Secondly, one can start by including all of the proposed pairwise feature combinations and progressively discard not-promising ones. Finally, for high-dimensional problems, one can do the opposite and progressively add pairwise feature combinations, starting from the empty set.

\subsection{Model Inference}
\label{subsec:inference}

Let $\vec{X}_{1:t} \in \R^{t \times D}$ be the matrix whose $i^{\text{th}}$ row contains the input $\x_i \in \X$ queried at iteration $i$, $\dim \X = D$, and let $\vec{y}_{1:t} = [y_1, \dots, y_t]^\top \in \mathbb{R}^t$ be the array of the corresponding noisy function observations.
Also, let $\vec{\Phi}_{1:t} \in \mathbb{R}^{t \times M}$ be the matrix whose $i^{\text{th}}$ row contains the featurized input $\ph(\x_i) \in \R^M$.
The formulation of $f$ in Eq.~\eqref{eq:f} and the noisy observation model induce the Gaussian likelihood
$p(\vec{y}_{1:t} | \vec{X}_{1:t}, \w) = \mathcal{N}(\vec{\Phi}_{1:t} \w, \beta^{-1} \vec{I})$.
To reflect our \emph{a priori} belief about the weight vector $\w$ and thus $f$, we specify a prior distribution over $\w$. A natural choice for this is a zero-mean isotropic Gaussian prior $p(\w|\alpha) = \mathcal{N}(\vec{0}, \alpha^{-1}\vec{I})$, with precision $\alpha > 0$, which encourages $\w$ to be uniformly small, so that the final predictor is a sum of all features, each giving a small, non-zero contribution.
Given the likelihood and prior, we infer the posterior 
$p(\w | \vec{X}_{1:t}, \vec{y}_{1:t}, \alpha, \beta) \propto p( \vec{y}_{1:t} | \vec{X}_{1:t}, \w, \beta) p(\w|\alpha)$, which due to conjugacy is Gaussian, $p(\w | \vec{X}_{1:t}, \vec{y}_{1:t}) = \mathcal{N}(\vec{m}, \vec{S}^{-1})$, with mean $\vec{m} = \beta\vec{S}^{-1} \vec{\Phi}_{1:t}^\top \vec{y}_{1:t} \in \mathbb{R}^M$ and precision $\vec{S} = \alpha \vec{I} + \beta\vec{\Phi}_{1:t}^\top \vec{\Phi}_{1:t} \in \mathbb{R}^{M \times M}$ \cite{williams2006gaussian}.
This simple analytical treatment of the posterior distribution over $\w$ is a main benefit of this model, which can be viewed as a GP with a linear kernel in feature space.

\subsection{Acquisition Function}
\label{subsec:acquisition}
We propose to use Thompson sampling (TS) \cite{thompson1933}, which samples weights $\widetilde{\w} \sim p(\w | \vec{X}_{1:t}, \vec{y}_{1:t}, \alpha, \beta)$ from the posterior and chooses the next input by solving $\widehat{\x} \in \operatorname{arg\ min}_{\x \in \X} \widetilde{\w}^\top \ph(\x)$.
TS intuitively focuses on inputs that are plausibly optimal and
has previously been successfully applied in both discrete and continuous domains \cite{baptista2018,mutny2018}.

TS requires solving $\widehat{\x} \in \operatorname{arg\ min}_{\x \in \X} \widetilde{\w}_t^\top \ph(\x)$, which is a challenging mixed-variable optimization problem.
However, as $\widetilde{\w}_t^\top \ph(\x)$ decomposes as in Eq.~\eqref{eq:f}, we can naturally use an alternating optimization scheme which iterates between optimizing the discrete variables $\x^d$ conditioned on a particular setting of the continuous variables $\x^c$ and vice versa, until convergence to some local optimum. 
While this scheme provides no theoretical guarantees, it is simple and thus widely and effectively applied in many contexts where the objective is hard to optimize.
In particular, we iteratively solve 
$\widehat{\x}^d \in \operatorname{arg\ min}_{\x^d \in \X^d} \big({{}\widetilde{\w}^d}^\top \ph^d(\x^d) + {{}\widetilde{\w}^m}^\top \ph^m(\x^d, \x^c = \widehat{\x}^c) \big)$,
$\widehat{\x}^c \in \operatorname{arg\ min}_{\x^c \in \X^c} \big({{}\widetilde{\w}^c}^\top \ph^c(\x^c) + {{}\widetilde{\w}^m}^\top \ph^m(\x^d = \widehat{\x}^d, \x^c)\big)$.
Importantly, using the mixed features proposed in Sec.~\ref{subsec:model}, these problems can be optimized by purely discrete and continuous optimizers, respectively. This also holds in the presence of mixed constraints $g^m(\x) \geq 0$ if those decompose accordingly into discrete and continuous constraints.

This scheme leverages independent subroutines for discrete and continuous optimization:
For the discrete part, we exploit the fact that optimizing a second-order pseudo-Boolean function is equivalent to a binary integer quadratic program (IQP) \cite{boros2002}, allowing us to exploit commonly-used efficient and robust solvers such as Gurobi or CPLEX.
While solving general binary IQPs is NP-hard \cite{boros2002}, these optimizers are in practice very efficient for the dimensionalities we consider (i.e., $D_d < 100$).
This approach allows us to use any functionality offered by these tools, such as the ability to optimize objectives subject to linear constraints $\vec{A} \x^d \leq \vec{b}$, $\vec{A} \in \R^{K \times D_d}, \vec{b} \in \R^K$ or quadratic constraints ${\x^{d}}^{\top} \vec{Q} \x^d + \vec{q}^\top \x^d \leq b$,  $\vec{Q} \in \R^{D_d \times D_d}, \vec{q} \in \R^{D_d}, b \in \R$.
For the continuous part, one can use optimizers commonly used in continuous BO, such as L-BFGS or DIRECT.
In our experiments, we use Gurobi as the discrete and L-BFGS
as the continuous solver within the alternating optimization scheme, which we always run until convergence.

\subsection{Model Discussion}
\label{subsec:discussion}
BO algorithms are comprised of three major design choices: the surrogate model to estimate the objective, the acquisition function to measure informativeness of the inputs and the acquisition function optimizer to select queries. Due to the widespread availability of general-purpose optimizers for continuous functions, continuous BO is mostly concerned with the first two design dimensions. However, this is different for mixed-variable constrained problems. We show in Sec.~\ref{sec:experiments} that using a heuristic optimizer for the acquisition function optimization leads to poor queries and, therefore, poor performance of the BO algorithm. Therefore, the tractability of the acquisition function optimization  influences and couples the other design dimensions.
In particular, the following considerations make the choice of a linear model and TS the ideal combination of surrogate and acquisition function for our problem.
Firstly, the linear model is preferable to a GP with a mixed-variable kernel as the latter would complicate the acquisition function optimization for two reasons: (i) the posterior samples would be arbitrary nonlinear functions of the discrete variables and (ii) it would be non-trivial to evaluate them at arbitrary points in the domain. In contrast, our explicit feature expansion solves both problems, while second order interactions provide a valid discrete function representation \cite{baptista2018,hazan2017} and lead to tractable quadratic MIPs with capacity for complex discrete constraints.
Moreover, Random Fourier Features approximate common GP kernels arbitrarily well, and inference in \textsc{MiVaBo} scales linearly with the number of data points, making it applicable in cases where GP inference, which scales cubically with the number of data points, would be prohibitive.
Secondly, TS induces a simple relation between the surrogate and the resulting optimization problem for the acquisition function, allowing to trade off model expressiveness and optimization tractability, which is a key challenge in mixed-variable domains.
Finally, the combination of TS and the linear surrogate facilitates the convergence analysis described in Sec.~\ref{subsec:regret_bounds}, making \textsc{MiVaBo} the first mixed-variable BO method with theoretical guarantees.

\subsection{Convergence Analysis}
\label{subsec:regret_bounds}
Using a linear model and Thompson sampling, we can leverage convergence analysis from linearly parameterized multi-armed bandits, a well-studied class of methods for solving structured decision making problems \cite{abeille2017}.
These also assume the objective to be linear in features $\ph(\x) \in \R^M$ with a fixed but unknown weight vector $\w \in \R^M$, i.e. $\mathbb{E}[f(\x)| \ph(\x)] = \w^\top \ph(\x)$, and aim to minimize the \emph{total regret} up to time $T$: $\mathcal R(T) = \sum_{t=1}^T (f(\x_*) - f(\x_t))$.
We obtain the following regret bound for \textsc{MiVaBo}:
\begin{proposition}
\label{th:regret}
Assume that the following assumptions hold in every iteration $t=1,\ldots,T$ of the \textsc{MiVaBo} algorithm:
\begin{enumerate}[wide, labelwidth=!, labelindent=0pt]
	\item $\widetilde{\w}_t \hspace{-0.5mm}\sim \hspace{-0.5mm}\mathcal{N}(\vec{m}, 24M\ln T\ln\frac{1}{\delta} \vec{S}^{-1})$, i.e.\ with scaled variance.
	\item $\x_t = \arg \min_{\x} \widetilde{\w}^\top \ph(\x)$ is selected exactly.\footnote{To this end, one can use more expensive but theoretically backed optimization methods instead of the alternating one, such as the powerful and popular \emph{dual decomposition} \cite{sontag2011}.}
    \item $\|\widetilde{\w}_t\|_2\leq c, \|\ph(\x_t)\|_2 \leq c, \|f(\x_*) \hspace{-0.5mm}- \hspace{-0.5mm}f(\x_t) \|_2 \leq c \ $, \hspace{-0.5mm}$c \in \mathbb R^+$\hspace{-1mm}.
\end{enumerate}
    Then, $\mathcal R(T) \leq \tilde{\mathcal{O}}\left(M^{3/2} \sqrt{T} \ln\frac{1}{\delta}\right)$ with probability $1- \delta$.
\end{proposition}
Prop.~\ref{th:regret} follows from Theorem 1 in \cite{abeille2017} and works for infinite arms $\x \in \X, |\X| = \infty$. In our setting, both the discrete and continuous Fourier features (and, thus, the mixed features) satisfy the standard boundedness assumption, such that the proof indeed holds. 
Prop.~\ref{th:regret} implies \emph{no-regret}, $\lim_{T \rightarrow \infty} \mathcal{R}(T) / T = 0$, i.e., convergence to the global minimum, since the minimum found after $T$ iterations is no further away from $f(\x_*)$ than the mean regret $\mathcal{R}(T) / T$.
To our knowledge, \textsc{MiVaBo} is the first mixed-variable BO algorithm for which such a guarantee is known to hold.

\section{Experiments}
We present experimental results on tuning the hyperparameters of two machine learning algorithms, namely gradient boosting and a deep generative model, on multiple datasets.

\label{sec:experiments}

\begin{figure*}[ht!]
    \begin{minipage}{0.7\textwidth}
        \vspace{-2mm}
        \hspace{-2mm}
        \begin{subfigure}[b]{0.5\textwidth}
            \includegraphics[width=1.05\textwidth]{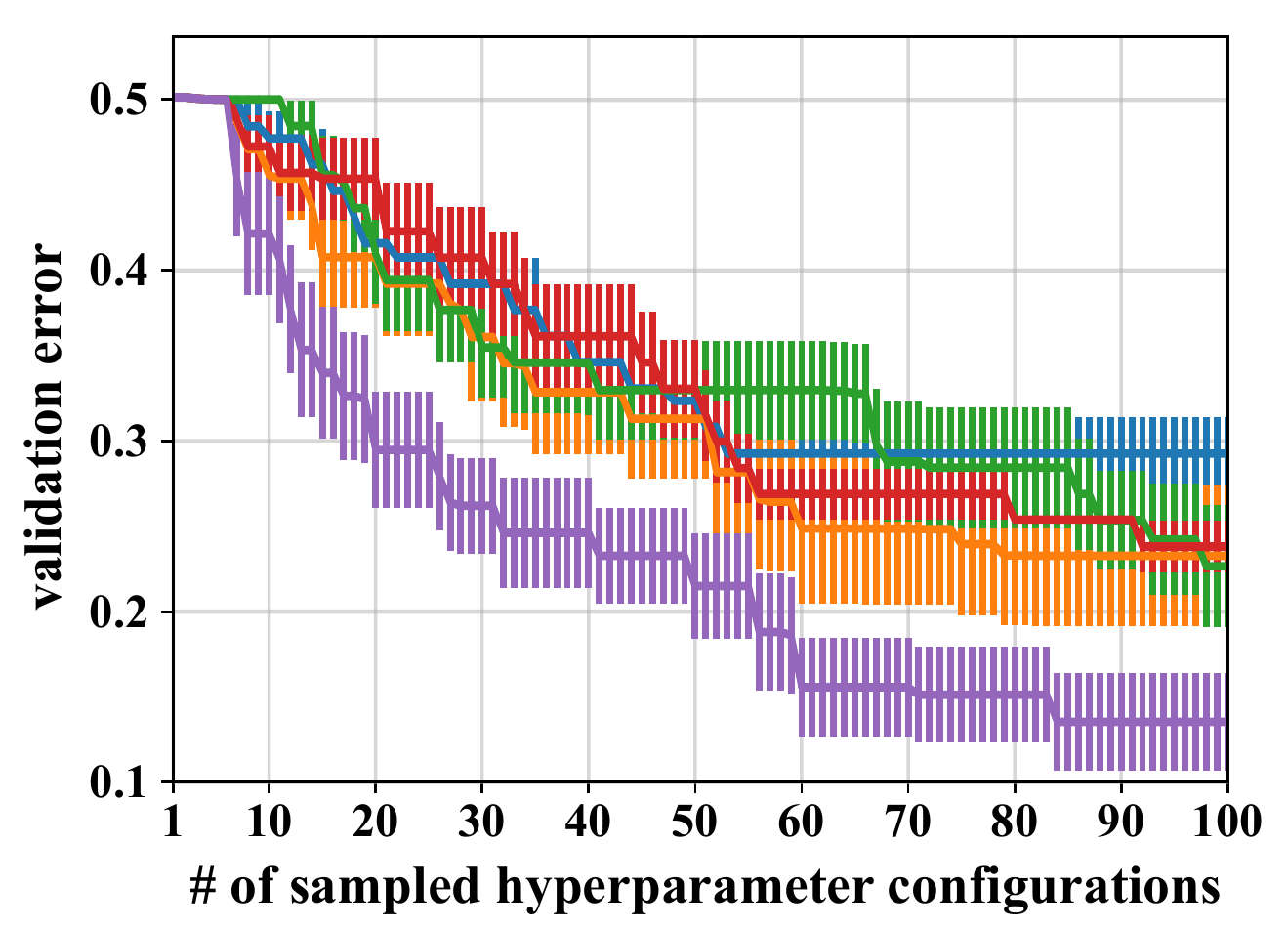}
        \end{subfigure}%
        \hspace{-1.5mm}
        \begin{subfigure}[b]{0.5\textwidth}
            \includegraphics[width=1.05\textwidth]{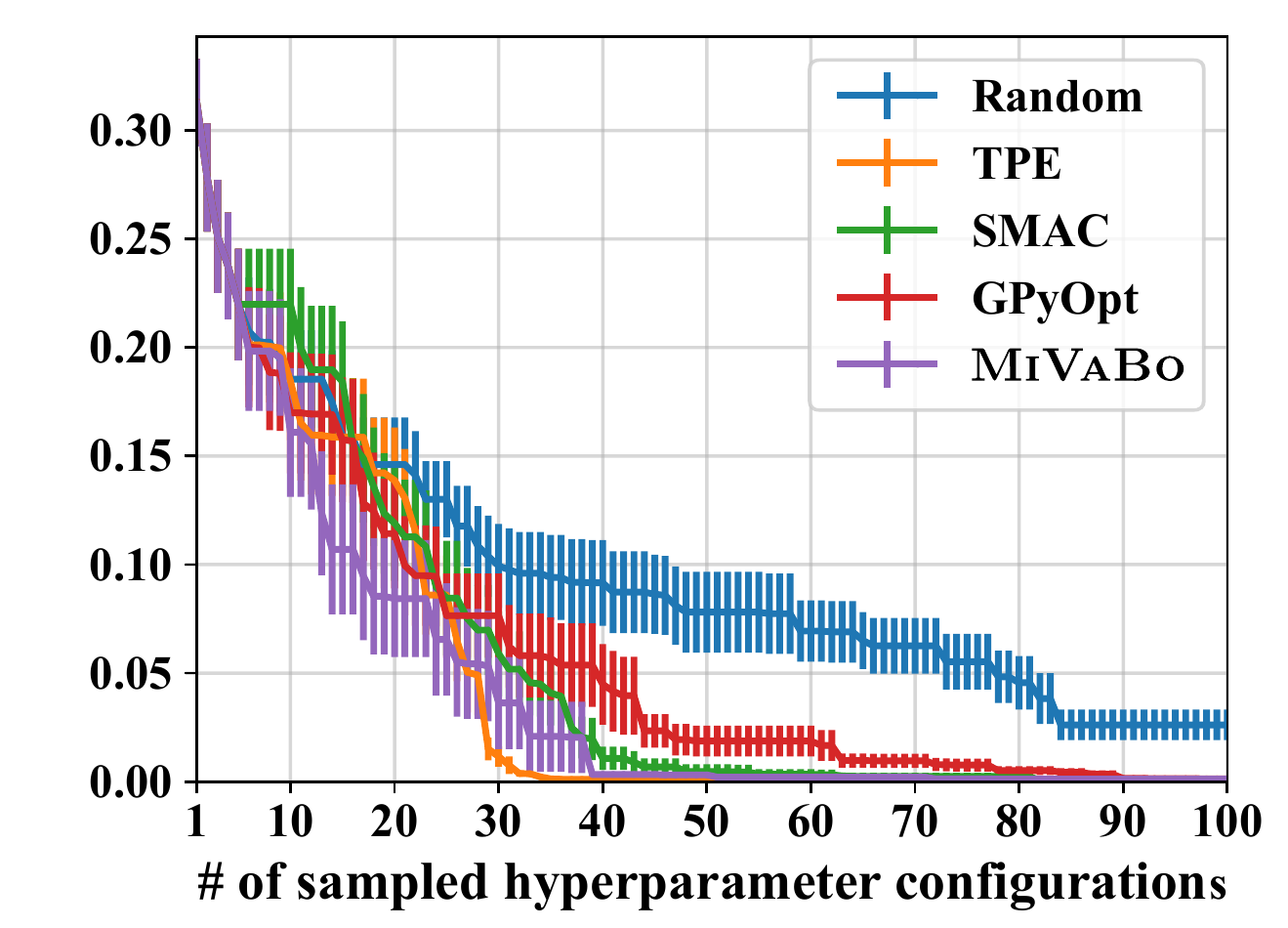}
        \end{subfigure}
        \vspace{-1.5mm}
        \caption{\textbf{XGBoost hyperparameter tuning} on \texttt{monks-problem-1} (left) and \texttt{steel-} \texttt{plates-fault} (right). Mean $\pm$ one std.\ of the validation error over 16 random seeds. \textsc{MiVaBo} significantly outperforms the baselines on the first dataset, and is competitive on the second.}
        \label{fig:xgboost}
    \end{minipage}\hfill
    \begin{minipage}{0.27\textwidth}
        \hspace{-2mm}
        \vspace{-2.5mm}
        \includegraphics[trim={0 7.0cm 0 5.0cm},clip,width=1.04\textwidth]{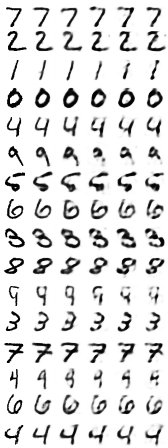}
        \captionof{figure}{Randomly chosen \texttt{MNIST} test images (left column) and their reconstructions by the best VAE models found by \textsc{MiVaBo}, random search, GPyOpt, TPE and SMAC (left to right), thus ordered by NLL values, which seem to capture visual quality.}
        \label{fig:reconstructions}
    \end{minipage}
\end{figure*}
\begin{figure*}[ht!]
    \begin{minipage}{0.7\textwidth}
        \hspace{-3mm}
        \vspace{-1.9mm}
        \begin{subfigure}[b]{0.5\textwidth}
            \includegraphics[width=1.05\textwidth]{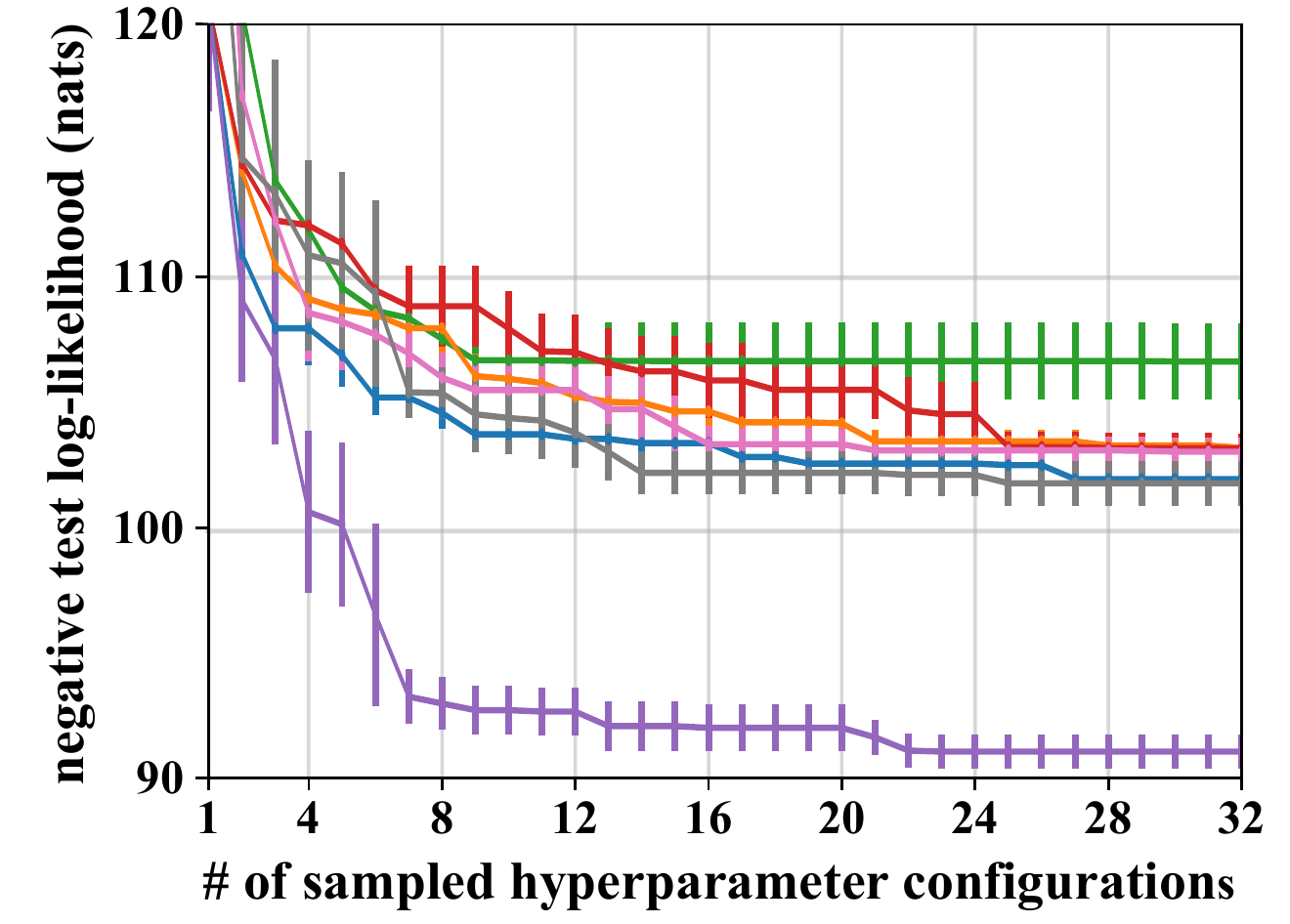}
        \end{subfigure}%
        \hspace{-1.5mm}
        \begin{subfigure}[b]{0.5\textwidth}
            \includegraphics[width=1.05\textwidth]{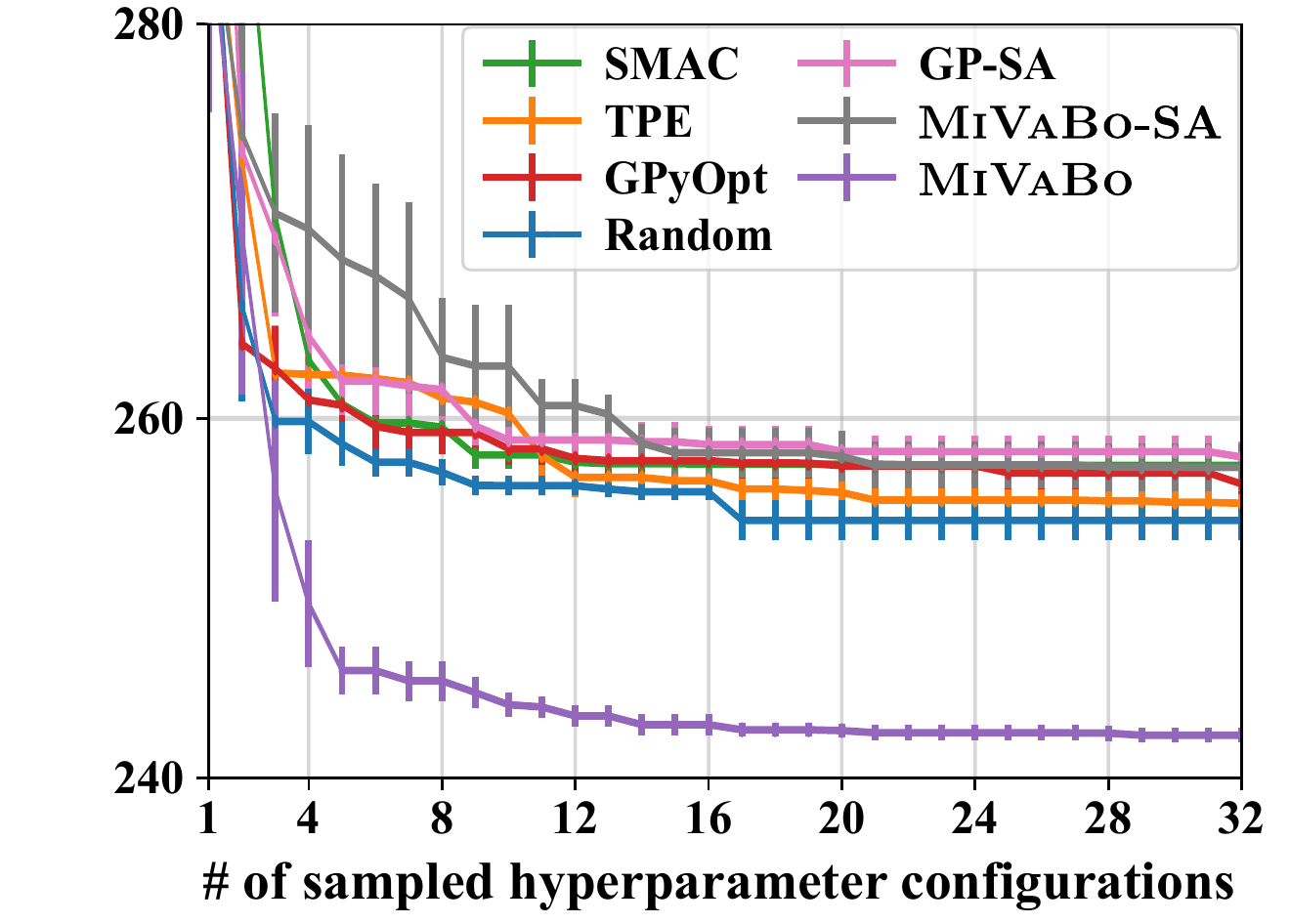}
        \end{subfigure}
        \caption{\textbf{VAE hyperparameter tuning} on \texttt{MNIST} (left) and \texttt{FashionMNIST} (right). Mean $\pm$ one std.\ of the NLL in nats, estimated using 32 importance samples, over 8 random seeds. Every model was trained for 32 epochs. \textsc{MiVaBo} significantly outperforms the state-of-the-art baselines, demonstrating its ability to handle the complex constrained nature of the VAE's parameter space.}
        \label{fig:vae_tuning}
    \end{minipage}\hfill
    \begin{minipage}{0.27\textwidth}
        \centering
        \begin{tabular}{ l r r }
            \toprule
            \textbf{Method} & \textbf{Time} & \textbf{NLL}\\
            \midrule
            \textbf{SMAC} & 0.32s & 99.09\\
            \textbf{TPE} & 0.12s & 97.05\\
            \textbf{GPyOpt} & 0.65s & 97.33\\
            \textbf{Random} & 0.01s & 93.74\\
            \textbf{\textsc{MiVaBo}} & 7.39s & \textbf{84.25}\\
            \bottomrule
        \end{tabular}
        \caption{Mean wall-clock time of one iteration (excluding function evaluation time) and mean negative log-likelihood (NLL) in nats, estimated with 5000 importance samples, of the best VAEs found after 32 BO iterations (as in \cref{fig:vae_tuning}), when trained for 3280 epochs. Human expert baseline for even deeper models is 82-83 nats.}
        \label{tab:nll}
    \end{minipage}
\end{figure*}

\paragraph{Experimental Setup.}
For \textsc{MiVaBo}\footnote{We provide a \texttt{Python} implementation of \textsc{MiVaBo} at \texttt{\url{https://github.com/edaxberger/mixed_variable_bo}}.}, we set the prior variance $\alpha$, observation noise variance $\beta$, and kernel bandwidth $\sigma$ to 1.0, and scale the variance as stated in Prop.~\ref{th:regret}.
We compare against SMAC, TPE, random search, and the popular GPyOpt BO package.
GPyOpt uses a GP model with the upper confidence bound acquisition function \cite{srinivas09}, and accounts for mixed variables by relaxing discrete variables to be continuous and later rounding them to the nearest discrete neighbor. 
To separate the influence of model choice and acquisition function optimization, we also consider the \textsc{MiVaBo} model optimized by simulated annealing (SA) %
(\textsc{MiVaBo}-SA) and the GP approach optimized by SA (GP-SA).
We compare against the SA-based variants only in constrained settings, using more principled methods in unconstrained ones.
To handle constraints, SA assigns high energy values to invalid inputs, making the probability of moving there negligible. 
We use SMAC, TPE and GPyOpt and SA with their respective default settings.

\subsection{Gradient Boosting Tuning}
The OpenML database \cite{vanschoren2014} contains evaluations for various machine learning methods trained on several datasets with many hyperparameter settings.
We consider extreme gradient boosting (XGBoost) \cite{chen2016}, one of the most popular OpenML benchmarks, and tune its ten hyperparameters -- three are discrete and seven continuous -- to minimize the classification error on a held-out test set (without any constraints).
We use two datasets, each containing more than $45000$ hyperparameter settings.
To evaluate hyperparameter settings for which no data is available, we use a surrogate modeling approach based on nearest neighbor \cite{eggensperger2015}, meaning that the objective returns the error of the closest (w.r.t.\ Euclidean distance) setting available in the dataset.
\cref{fig:xgboost} shows that \textsc{MiVaBo} achieves performance which is either significantly stronger than (left dataset) or competitive with (right dataset) the state-of-the-art mixed-variable BO algorithms on this challenging task.
GPyOpt performs poorly, likely because it cannot account for discrete variables in a principled way.
As compared to TPE and SMAC, \textsc{MiVaBo} seems to benefit from more sophisticated uncertainty estimation.

\subsection{Deep Generative Model (DGM) Tuning}
DGMs recently received considerable attention in the machine learning community. Despite their popularity and importance, effectively tuning their hyperparameters is a major challenge.
We consider tuning the hyperparameters of a variational autoencoder (VAE) \cite{kingma2013} composed of a convolutional encoder and a deconvolutional decoder \cite{salimans2014}.
The VAEs are evaluated on stochastically binarized \texttt{MNIST}, as in \cite{burda2015}, and \texttt{FashionMNIST}.
They are trained on 60000 images for 32 epochs, using Adam with a mini-batch size of 128.
We report the negative log-likelihood (NLL; in nats) achieved by the VAEs on a held-out test set of 10000 images, as estimated via importance sampling using 32 samples per test point.
To our knowledge, no other BO paper considered DGM tuning.

VAE tuning is difficult due to the high-dimensional and structured nature of its hyperparameter space, and, in particular, due to constraints arising from dependencies between some of its parameters.
We tune 25 discrete parameters defining the model architecture, e.g. the number of convolutional layers, their stride, padding and filter size, the number and width of fully-connected layers, and the latent space dimensionality.
We further tune three continuous parameters for the optimizer and regularization.
Crucially, mutual dependencies between the discrete parameters result in complex constraints, as certain combinations of stride, padding and filter size lead to invalid architectures. Particularly, for the encoder, the shapes of all layers must be integral, and for the decoder, the output shape must match the input data shape, i.e., one channel of size $28 \times 28$ for \texttt{\{Fashion\}MNIST}.
The latter constraint is especially challenging, as only a small number of decoder configurations yield the required output shape.
Thus, even for rather simple datasets such as \texttt{\{Fashion\}MNIST}, tuning such a VAE is significantly more challenging than, say, tuning a convolutional neural network for classification.

While \textsc{MiVaBo} can conveniently capture these restrictions via linear and quadratic constraints, the competing methods cannot.
To enable a comparison that is as fair as possible, we thus use the following sensible heuristic to incorporate the knowledge about the constraints into the baselines: If a method tries to evaluate an invalid parameter configuration, we return a penalty error value, which will discourage a model-based method to sample this (or a similar) setting again. However, for fairness, we only report valid observations and ignore all configurations that violated a constraint.
We set the penalty value to 500 nats, which is the error incurred by a uniformly random generator.
We investigated the impact of the penalty value (e.g., we also tried 250 and 125 nats) and found that it does \emph{not} qualitatively affect the results.

\cref{fig:vae_tuning} shows that \textsc{MiVaBo} significantly outperforms the competing methods on this task, both on \texttt{MNIST} (left) and \texttt{FashionMNIST} (right).
This is because \textsc{MiVaBo} can naturally encode the constraints and thus directly optimize over the feasible region in parameter space, while TPE, SMAC and GPyOpt need to learn the constraints from data.
They fail to do so and get stuck in bad local optima early on.
The model-based approaches likely struggle due to sharp discontinuities in hyperparameter space induced by the constraint violation penalties (i.e., as invalid configurations may lie close to well-performing configurations).
In contrast, random search is agnostic to these discontinuities, and thus notably outperforms the model-based methods.
Lastly, GP-SA and \textsc{MiVaBo}-SA struggle as well, suggesting that while SA can avoid invalid inputs, the effective optimization of complex constrained objectives crucially requires more principled approaches for acquisition function optimization, such as the one we propose.
This shows that all model choices for \textsc{MiVaBo} (as discussed in Sec.~\ref{subsec:discussion}) are necessary to achieve such strong results.

Although log-likelihood scores allow for a quantitative comparison, they are hard to interpret for humans.
Thus, for a qualitative comparison, \cref{fig:reconstructions} visualizes the reconstruction quality achieved on \texttt{MNIST} by the best VAE configuration found by all methods after 32 BO iterations.
The VAEs were trained for 32 epochs each, as in \cref{fig:vae_tuning}.
The likelihoods seem to correlate with the quality of appearance, and the model found by \textsc{MiVaBo} arguably produces the visually most appealing reconstructions among all models.
Note that while \textsc{MiVaBo} requires more time than the baselines (see \cref{tab:nll}), this is still negligible compared to the cost of a function evaluation, which involves training a deep generative model.
Finally, the best VAE found by \textsc{MiVaBo} achieves $84.25$ nats on \texttt{MNIST} when trained for 3280 epochs and using 5000 importance samples for log-likelihood estimation, i.e.\ the setting used in \cite{burda2015} (see \cref{tab:nll}).
This is comparable to the performance of 82-83 nats achieved by human expert tuned models, e.g.\ as reported in \cite{salimans2014} (which use even more convolutional layers and a more sophisticated inference method), highlighting \textsc{MiVaBo}'s effectiveness in tuning complex deep neural network architectures.

\section{Conclusion}
\label{sec:conclusion}
We propose \textsc{MiVaBo}, the first method for efficiently optimizing expensive mixed-variable black-box functions subject to linear and quadratic discrete constraints.
\textsc{MiVaBo} combines a linear model of expressive features with Thompson sampling, making it simple yet effective. Moreover, it is highly flexible due to the modularity of its components, i.e., the mixed-variable features, and the optimization oracles for the acquisition procedure. This allows practitioners to tailor \textsc{MiVaBo} to specific objectives, e.g. by incorporating prior knowledge in the feature design or by leveraging optimizers handling specific types of constraints. We show that \textsc{MiVaBo} enjoys theoretical convergence guarantees that competing methods lack. Finally, we empirically demonstrate that \textsc{MiVaBo} significantly improves optimization performance as compared to state-of-the-art methods for mixed-variable optimization on complex hyperparameter tuning tasks.

\section*{Acknowledgements}
This research has been partially supported by SNSF NFP75 grant 407540\_167189.
Matteo Turchetta was supported through the ETH-MPI Center for Learning Systems.
Erik Daxberger was supported through the EPSRC and Qualcomm.
The authors thank Josip Djolonga, Mojm\'{i}r Mutn\'{y}, Johannes Kirschner, Alonso Marco Valle, David R. Burt, Ross Clarke, Wolfgang Roth as well as the anonymous reviewers of an earlier version of this paper for their helpful feedback.

\putbib
\end{bibunit}

\appendix
\clearpage
\begin{bibunit}
\section{Sparse Linear Model via Sparsity-Encouraging Prior}
\label{sec:sparse_linear_models}
While the number and degree of the features used in the surrogate model (see \cref{subsec:model}) is a design choice, in practice it is typically unknown which variable interactions matter and thus which features to choose.
To discard irrelevant features, one may impose a sparsity-encouraging prior over the weight vector $\w$ \cite{baptista2018}.
However, due to non-conjugacy to the Gaussian likelihood, exact Bayesian inference of the resulting posterior distribution is in general intractable, imposing the need for approximate inference methods.
One choice for such a prior is the Laplace distribution, i.e. $p(\w|\alpha) \propto \exp(-\alpha^{-1}\|\w\|_1)$, with inverse scale parameter $\alpha > 0$, for which approximate inference techniques based on expectation propagation \cite{minka2001} and variational inference \cite{wainwright2008} were developed in \cite{seeger2008a,seeger2008b,seeger2011}.
Alternatively, one can use a horseshoe prior and use Gibbs sampling to sample from the posterior over weights \cite{baptista2018}. However, this comes with a significantly larger computational burden, which is a well-known issue for sampling based inference techniques \cite{bishop2006}.
Lastly, one may consider a spike-and-slab prior with expectation propagation for approximate posterior inference \cite{hernandez2013generalized,hernandez2015expectation}.

\section{Pseudocode for Thompson Sampling}
\label{sec:pseudo}
\cref{alg:mivabo} shows pseudocode for the Thompson sampling procedure within \textsc{MiVaBo}.
\begin{algorithm}[ht!]
\centering
\caption{\textsc{Thompson Sampling}}
\label{alg:mivabo}
\begin{algorithmic}[1]
    \REQUIRE model features $\ph(\x)$
    \STATE Set $\vec{S} = \vec{I}$, $\vec{m} = \vec{0}$
	\FOR{$t = 1, 2, \hdots, T$}
	    \STATE Sample $\widetilde{\w}_t \sim \mathcal{N}(\vec{m}, \vec{S}^{-1})$
        \STATE Select input $\widehat{\x}_t \in \operatorname{arg\,min}_{\x \in \X} \widetilde{\w}_t^\top \ph(\x)$
        \STATE Query output $y_t = f(\widehat{\x}_t) + \epsilon$
        \STATE Update $\vec{S}, \vec{m}$ as described in \cref{subsec:inference}.
    \ENDFOR
	\STATE {\bfseries Output:} $\widehat{\x}_* \in \operatorname*{arg\,min}_{\x \in \X} \vec{m}^\top \ph(\x)$
\end{algorithmic}
\end{algorithm}

\section{Further Experimental Results}
\label{sec:further_experiments}

\subsection{Synthetic Benchmark for Unconstrained Optimization}
\label{subsec:synthetic_unconstrained}
\begin{figure*}[ht!]
    \begin{tabular}{cc}
        \hspace{-4mm}
        \includegraphics[width=0.5\textwidth]{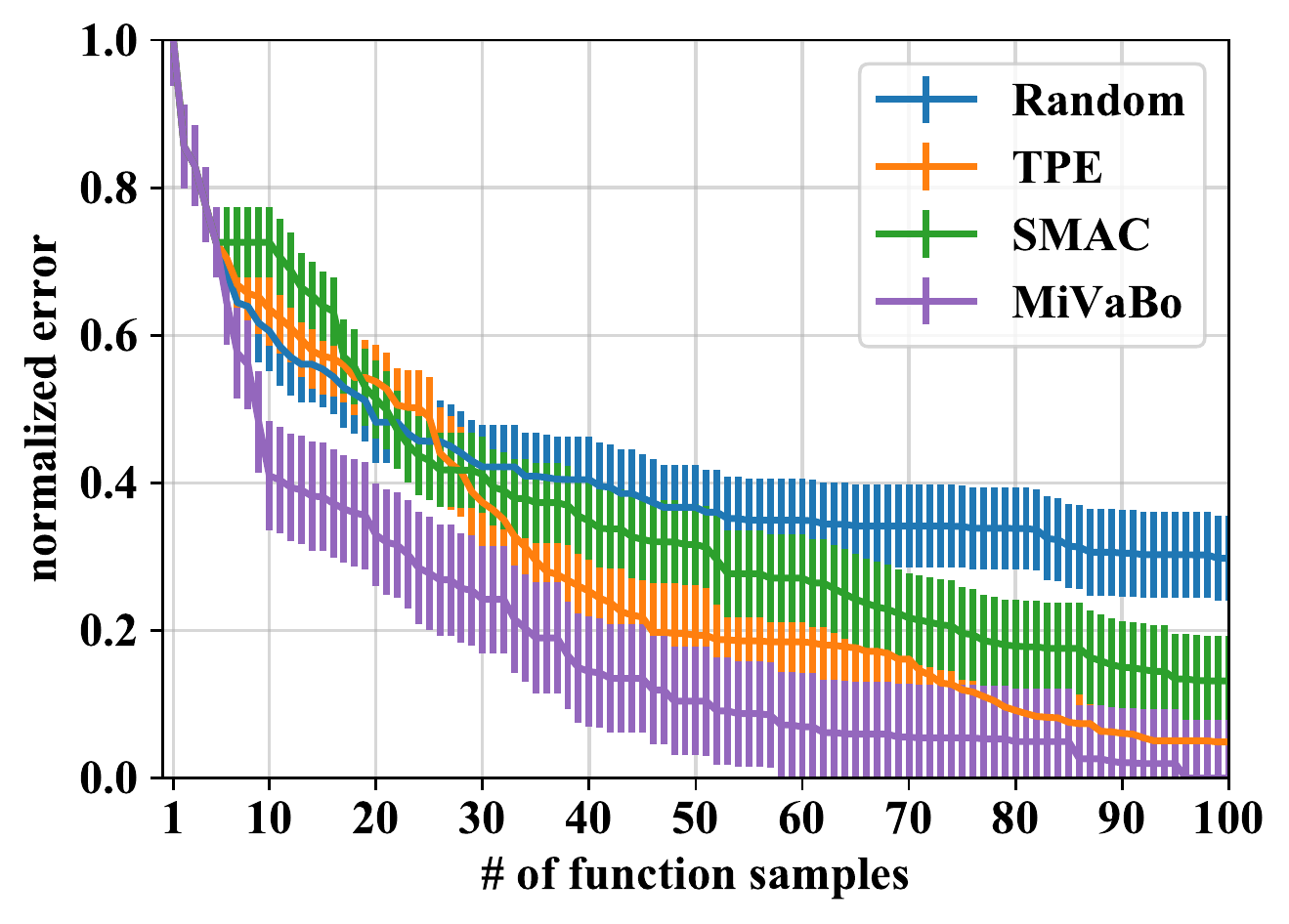}
        \includegraphics[width=0.5\textwidth]{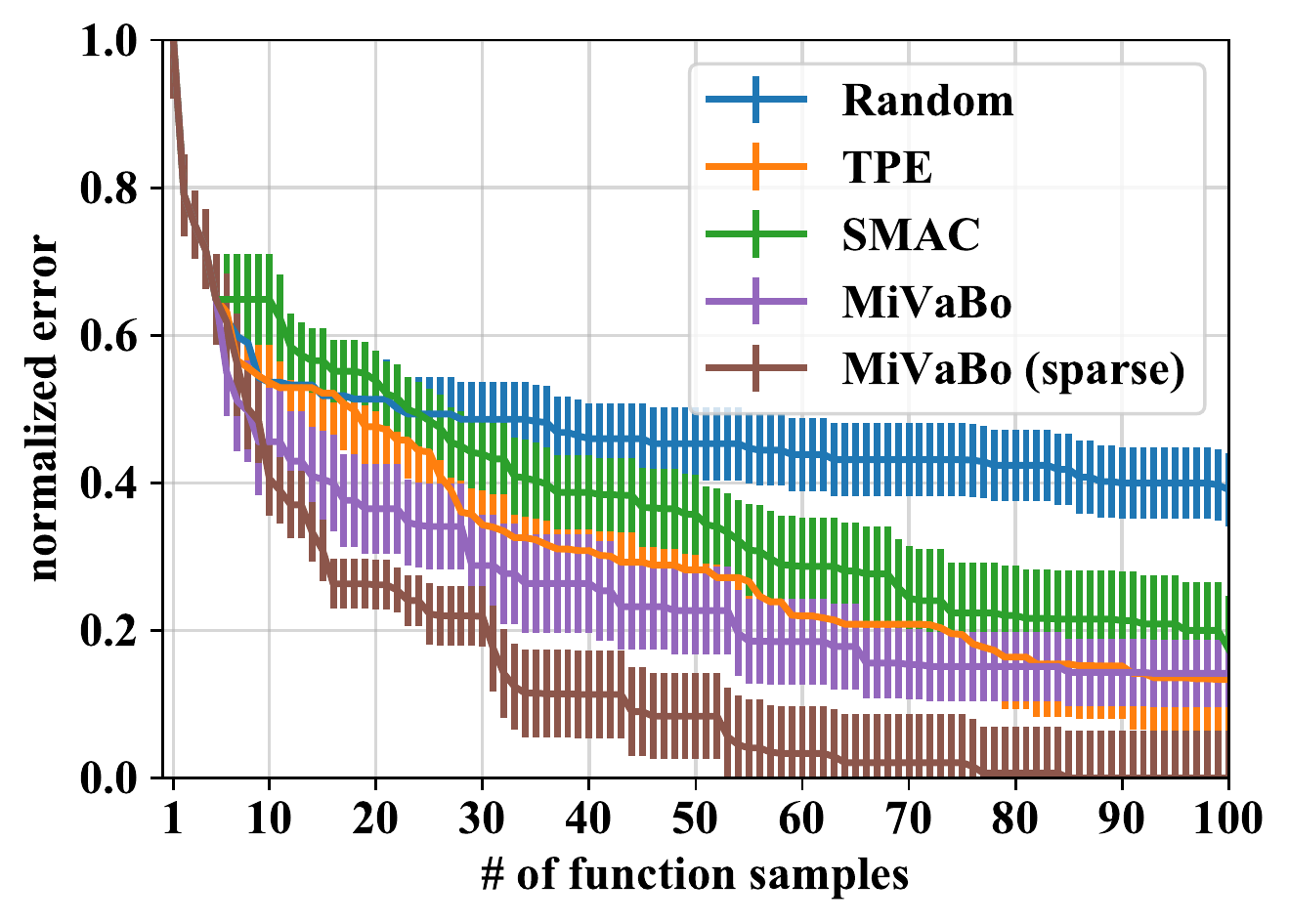}
    \end{tabular}
    \caption{Results on the synthetic benchmark, with the Gaussian (left) and Laplace prior (right). Mean plus/minus one standard deviation of the normalized error over 16 random initializations. (Left) \textsc{MiVaBo} outperforms its competitors. (Right) \textsc{MiVaBo} with a sparse prior outperforms its competitors, including \textsc{MiVaBo} with a Gaussian prior}
    \label{fig:synthetic}
\end{figure*}
We assess the performance on an unconstrained synthetic linear benchmark function of the form $f(\x) = \w^\top \ph$.
We choose a fairly high-dimensional objective with $D_d = 8$ discrete and $D_c = 8$ continuous variables, thus resulting in a total input space dimensionality of $D = D_d + D_c = 16$.
For the discrete model part, we choose the $M_d \in \mathcal{O}(D_d^2)$ features $\ph^d$ proposed in \cref{subsec:model}.
For the continuous features $\ph^c$, we choose $M_c = 16$ dimensional Random Fourier Features to approximate a GP with a squared exponential kernel with bandwidth $\sigma = 1.0$.
For the mixed representation, we construct a feature vector $\ph^m$ by stacking all pairs of discrete and continuous features, as proposed in \cref{subsec:model}.
We consider two settings for the weight vector:
Firstly, we sample it from a zero-mean Gaussian, $\w \sim \mathcal{N}(\vec{0}, \vec{I}) \in \R^M$.
Secondly, we sample it from a Laplace distribution, i.e. $\w \sim p(\w|\alpha) \propto \exp(-\alpha^{-1}\|\w\|_1)$, with inverse scale parameter $\alpha = 0.1$, and then prune all weights smaller than $10$ to zero to induce sparsity over the weight vector.
For the second setting, we also assess \textsc{MiVaBo} using a Laplace prior and the approximate inference technique from \cite{seeger2011} (see also \cref{sec:sparse_linear_models})\footnote{We use the MATLAB implementation provided in the \texttt{glm-ie} toolbox \cite{nickisch2012} by the same authors.}.
As we do not know the true optimum of the function and thus cannot compute the regret, we normalize all observed function values to the interval $[0,1]$, resulting in a normalized error as the metric of comparison.
We can observe from our results shown in \cref{fig:synthetic} that \textsc{MiVaBo} outperforms the competing methods in this setting, demonstrating the effectiveness of our approach when its modeling assumptions are fulfilled. 

\subsection{Synthetic Benchmark for Constrained Optimization}
\label{subsec:synthetic_constrained}
In another experiment, we demonstrate the capability of our algorithm to incorporate linear constraints on the discrete variables.
In particular, we want to enforce a solution that is sparse in the discrete variables via adding a hard cardinality constraint of the type $\sum_{i=1}^{D_d} x^d_i \leq k$, which we can simply specify in the Gurobi optimizer.
Cardinality constraints of this type are very relevant in practice, as many real-world problems desire sparse solutions (e.g., sparsification of ising models, contamination control, aero-structural multi-component problems \cite{baptista2018}).
We consider the same functional form as before, i.e. again with $D_d = D_c = 8$, and set $k = 2$, meaning that a solution should have at most two of our binary variables set to one, while all others shall be set to zero.
To enable comparison with TPE, SMAC and random search, which provide no capability of modeling these kinds of constraints, we assume the objective $f$ to be unconstrained, but instead return a large penalty value if a method acquires an evaluation of $f$ at a point that violates the constraint.
Thus, the baseline algorithms are forced to learn the constraint from observations, which is a challenging problem.

\begin{figure}[ht!]
\centering
    \includegraphics[width=0.5\textwidth]{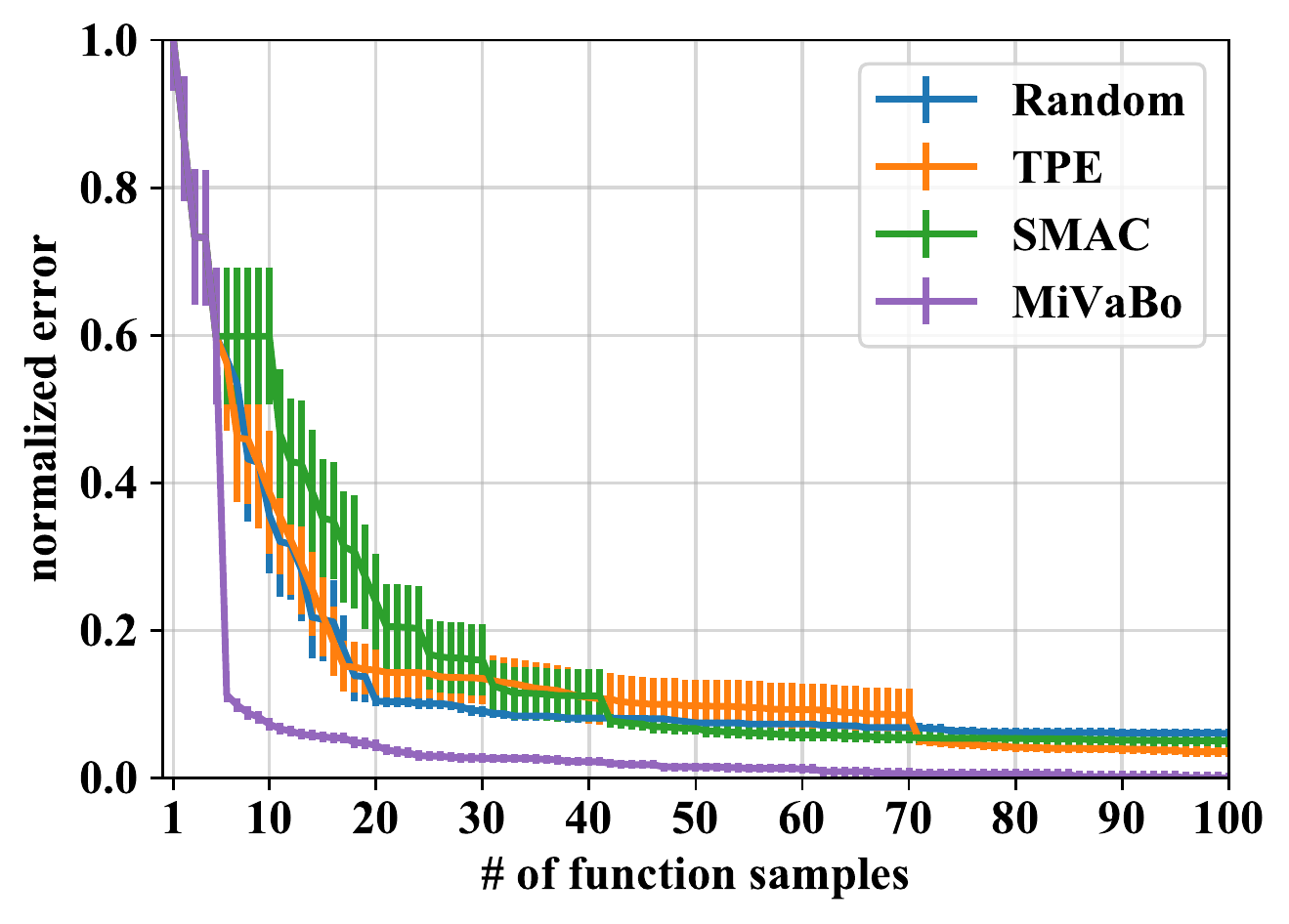}
    \caption{Results on the synthetic benchmark with cardinality constraints. The curves represent the mean plus/minus one standard deviation of the normalized error over 16 random initializations. One can observe that \textsc{MiVaBo} outperforms its competitors.}
    \label{fig:constraint}
\end{figure}

One can notice from \cref{fig:constraint} that the ability to explicitly encode the cardinality constraint into the discrete optimization oracle significantly increases performance.

\section{More Details on XGBoost Hyperparameter Tuning Task}
\label{sec:xgboost}
Please refer to the corresponding websites for details on the OpenML XGBoost benchmark (\url{https://www.openml.org/f/6767}), on the underlying implementation (\url{https://www.rdocumentation.org/packages/xgboost/versions/0.6-4}), and on the \texttt{steel-plates-fault} (\url{https://www.openml.org/t/9967}) and \texttt{monks-problem-1} (\url{https://www.openml.org/t/146064}) datasets.
Finally, see \cref{tab:hypers_xgboost} for a description of the hyperparameters involved in XGBoost.
\begin{table}[ht!]
    \centering
    \caption{Hyperparameters of the XGBoost algorithm. 10 parameters, 7 of which are continuous, and 3 of which are discrete. }
    \begin{tabular}{ l l l }
        \toprule
        \textbf{Name} & \textbf{Type} & \textbf{Domain}\\
        \midrule
        booster & discr. & ['gbtree', 'gblinear']\\
        nrounds & discr. & $[3, 5000]$\\
        alpha & contin. & $[0.000985, 1009.209690]$\\
        lambda & contin. & $[0.000978, 999.020893]$\\
        colsample\_bylevel & contin. & $[0.046776, 0.998424]$\\
        colsample\_bytree & contin. & $[0.062528, 0.999640]$\\
        eta & contin. & $[0.000979, 0.995686]$\\
        max\_depth & discr. & $[1, 15]$\\
        min\_child\_weight & contin. & $[1.012169, 127.041806]$\\
        subsample & contin. & $[0.100215, 0.999830]$\\
        \bottomrule
    \end{tabular}
    \label{tab:hypers_xgboost}
\end{table}

\section{More Details on VAE Hyperparameter Tuning Task}
\label{sec:details_vae}
\subsection{Hyperparameters of VAE}
\label{subsec:hypers_vae}
We used the \texttt{PyTorch} library to implement the VAE used in the experiment.
\cref{tab:hypers_vae} describes the names, types and domains of the involved hyperparameters that we tune.
Whenever we refer to a "deconvolutional layer" (also called transposed convolution or fractionally-strided convolution), we mean the functional mapping implemented by a \texttt{ConvTranspose2d} layer in \texttt{PyTorch}\footnote{See \url{https://pytorch.org/docs/stable/nn.html\#convtranspose2d} for details.}. 
Since our approach operates on a binary encoding of the discrete parameters, we also display the number of bits required to encode each discrete parameter.
In total, we consider 25 discrete parameters (resulting in 50 when binarized) as well as three continuous ones.
\begin{table*}[ht!]
    \centering
    \caption{\textbf{Hyperparameters of the VAE.} The architecture of the VAE (if all layers are enabled) is \texttt{C1-C2-F1-F2-z-F3-F4-D1-D2}, with \texttt{C} denoting a convolutional (conv.) layer, \texttt{F} a fully-connected (fc.) layer, \texttt{D} a deconvolutional (deconv.) layer and \texttt{z} the latent space. Layers \texttt{F2} and \texttt{F3} have fixed sizes of $2 d_z$ and $d_z$ units respectively, where $d_z$ denotes the dimensionality of the latent space \texttt{z}. The domain of the number of units of the fc. layers \texttt{F1} and \texttt{F4} is discretized with a step size of 64, i.e. $[0,64,128,\ldots,832,896,960]$, denoted by $[0\ldots960]$ in the table for brevity. For $d_z$, the domain $[16\ldots64]$ refers to all integers within that interval.}
    \begin{tabular}{ r l l l l }
        \toprule
        \textbf{\#} & \textbf{Name} & \textbf{Type} & \textbf{Domain} & \textbf{Bits}\\
        \midrule
        1 & Number of conv. layers in encoder & discrete & [0,1,2] & 2\\
        & Parameters of \texttt{C1} & & \\
        2 & \hspace{5mm} Number of channels of \texttt{C1} & discrete & [4,8,16,24] & 2\\
        3 & \hspace{5mm} Stride of \texttt{C1} & discrete & [1,2] & 1\\
        4 & \hspace{5mm} Filter size of \texttt{C1} & discrete & [3,5] & 1\\
        5 & \hspace{5mm} Padding of \texttt{C1} & discrete & [0,1,2,3] & 2\\
        & Parameters of \texttt{C2} & & \\
        6 & \hspace{5mm} Number of channels of \texttt{C2} & discrete & [8,16,32,48] & 2\\
        7 & \hspace{5mm} Stride of \texttt{C2} & discrete & [1,2] & 1\\
        8 & \hspace{5mm} Filter size of \texttt{C2} & discrete & [3,5] & 1\\
        9 & \hspace{5mm} Padding of \texttt{C2} & discrete & [0,1,2,3] & 2\\
        10 & Number of fc. layers in encoder & discrete & [0,1,2] & 2\\
        11 & \hspace{5mm} Number of units of \texttt{F1} & discrete & [0\ldots960] & 4\\
        12 & Dimensionality $d_z$ of \texttt{z} & discrete & [16\ldots64] & 6\\
        13 & Number of fc. layers in decoder & discrete & [0,1,2] & 2\\
        14 & \hspace{5mm} Number of units of \texttt{F4} & discrete & [0\ldots960] & 4\\
        15 & Number of deconv. layers in decoder & discrete & [0,1,2] & 2\\
        & Parameters of \texttt{D1} & & \\
        16 & \hspace{5mm} Number of channels of \texttt{D1} & discrete & [8,16,32,48] & 2\\
        17 & \hspace{5mm} Stride of \texttt{D1} & discrete & [1,2] & 1\\
        18 & \hspace{5mm} Filter size of \texttt{D1} & discrete & [3,5] & 1\\
        19 & \hspace{5mm} Padding of \texttt{D1} & discrete & [0,1,2,3] & 2\\
        20 & \hspace{5mm} Output padding of \texttt{D1} & discrete & [0,1,2,3] & 2\\
        & Parameters of \texttt{D2} & & \\
        21 & \hspace{5mm} Number of channels of \texttt{D2} & discrete & [4,8,16,24] & 2\\
        22 & \hspace{5mm} Stride of \texttt{D2} & discrete & [1,2] & 1\\
        23 & \hspace{5mm} Filter size of \texttt{D2} & discrete & [3,5] & 1\\
        24 & \hspace{5mm} Padding of \texttt{D2} & discrete & [0,1,2,3] & 2\\
        25 & \hspace{5mm} Output padding of \texttt{D2} & discrete & [0,1,2,3] & 2\\
        \midrule
        26 & Learning rate & continuous & $[10^{-4}, 10^{-2}]$ & -\\
        27 & Learning rate decay factor & continuous & $[0.5,1.0]$ & -\\
        28 & Weight decay regularization & continuous & $[10^{-6},10^{-2}]$ & -\\
        \midrule
        & \textbf{Total} & & & \textbf{50}\\
        \bottomrule
    \end{tabular}
    \label{tab:hypers_vae}
\end{table*}

\subsection{Description of Constraints}
\label{subsec:constraints_vae}
We now describe the constraints arising from the mutual dependencies within the hyperparameter space of the deconvolutional VAE (as described in \cref{subsec:hypers_vae}).

\paragraph{Encoder constraints.}
For the convolutional layers (up to two in our case) of the encoder, we need to ensure that the chosen combination of stride, padding and filter size transforms the input image into an output image whose shape is integral (i.e., not fractional). More precisely, denoting the input image size by $W_\text{in}$ (i.e., the input image is quadratic with shape $W_\text{in} \times W_\text{in}$), the stride by $S$, the filter size by $F$, and the padding by $P$, we need to ensure that the output image size $W_\text{out}$ is integral, i.e.
\begin{equation}
\label{eq:w_out}
    W^e_\text{out} = (W^e_\text{in} - F^e + P^e) / S^e + 1 \in \mathbb{N}
\end{equation}
where superscripts $e$ are used to make clear that we are considering the encoder.
Let us illustrate this with an example\footnote{This example is taken from \url{http://cs231n.github.io/convolutional-networks/\#conv} (paragraph "Constraints on strides"), which also describes the constraints discussed here. Note that they define the padding $P$ in a slightly different way (i.e., they only consider symmetric padding, while we also allow for asymmetric padding) and thus end up with a term of $2P$ instead of $P$ in the formula.}:
For $W_\text{in} = 10$, $P = 0$, $S = 2$ and $F = 3$, we would get an invalid fractional output size of $W_\text{out} = (10 - 3 + 0) / 2 + 1 = 4.5$.
To obtain a valid output size, one could, e.g., instead consider a padding of $P=1$, yielding $W_\text{out} = (10 - 3 + 1) / 2 + 1 = 5$.
Alternatively, one could also consider a stride of $S = 1$ to obtain $W_\text{out} = (10 - 3 + 0) / 1 + 1 = 8$, or a filter size of $F = 4$ to obtain $W_\text{out} = (10 - 4 + 0) / 2 + 1 = 4$ (though the latter is very uncommon and thus not allowed in our setting; we only allow $F \in \{3,5\}$, as described in \cref{subsec:hypers_vae}).
While this constraint is not trivially fulfilled (which can be verified by manually trying different configurations of $W_\text{in}, F, S, P$), it is also not too challenging to find valid configurations.

Note that this constraint is required to be fulfilled for every convolutional layer; we thus obtain the following two constraints in our specific two-layer setting, where $W_\text{in} = 28$ (as \texttt{MNIST} and \texttt{FashionMNIST} images are of shape $28 \times 28$):
\begin{align}
    &W^e_\text{out1} = (28 - F^e_1 + P^e_1) / S^e_1 + 1&\in \mathbb{N} \label{eq:const_e_1},\\
    &W^e_\text{out2} = (W^e_\text{out1} - F^e_2 + P^e_2) / S^e_2 + 1&\in \mathbb{N}. \label{eq:const_e_2}
\end{align}
where the subscripts in $\{1,2\}$ denote the index of the convolutional layer.

Finally, observe that the constraints in Eq.~\eqref{eq:const_e_1} and Eq.~\eqref{eq:const_e_2} are, respectively, linear and quadratic in the discrete variables $F^e_1, F^e_2, P^e_1, P^e_2, S^e_1, S^e_2$, and can thus be readily incorporated into the integer programming solver (e.g. Gurobi \cite{optimization2014} or CPLEX \cite{cplex2009v12}) we employ as a subroutine within our acquisition function optimization strategy.

\paragraph{Decoder constraints.}
While the constraints on the decoder architecture are similar in nature to those for the encoder, they are significantly more difficult to fulfill, which we will now illustrate.

In particular, we need to ensure that the decoder produces images of shape $28 \times 28$. By inverting the formula in Eq.~\eqref{eq:w_out}, we see that for a deconvolutional layer (which intuitively implements an inversion of the convolution operation), the output image size $W_\text{out}$ can be computed as
\begin{align}
\label{eq:w_out_d}
    W^d_\text{out} = (W^d_\text{in} - 1) \times S^d + F^d - 2 P^d + O^d
\end{align}
where superscripts $d$ are used to make clear that we are considering the decoder, and where $O$ is an additional output padding parameter which can be used to adjust the shape of the output image\footnote{See e.g. \url{https://pytorch.org/docs/stable/nn.html\#convtranspose2d} for a description of the output padding in the context of the \texttt{PyTorch} library we use.}.
Note that we now have a factor of $2 P$ in Eq.~\eqref{eq:w_out_d} instead of $P$ (as for the encoder, i.e. in Eq.~\eqref{eq:w_out}), since we only consider symmetric padding for the decoder, while we allow for asymmetric padding for the encoder (to make it easier to fulfill the integrality constraints for the encoder due to an increased number of valid configurations).
The output padding parameter $O$ is required since the mapping from $W^e_\text{in}$ to $W^e_\text{out}$ in a convolutional layer (i.e. in the encoder) is not bijective: there are different combinations of $W^e_\text{in}, F, S, P$ that result in the same $W^e_\text{out}$ (which can be easily verified).
Thus, given an output size $W^e_\text{out}$ (now serving as the input size $W^d_\text{in}$ of the deconvolutional layer), there is no unique corresponding input size $W^e_\text{in}$ (now serving as the output size $W^d_\text{out}$ of the deconvolutional layer).
The output padding parameter $O$ can thus be used to disambiguate this relation.
Note that $W^d_\text{out}$ in Eq.~\eqref{eq:w_out_d} is always integral, so there are no integrality constraints involved here, in constrast to the encoder.

In the context of our decoder model, i.e. with up to two deconvolutional layers, and with a required output image size of $28$, we thus obtain the following constraints:
\begin{align}
    W^d_\text{out} = (W^d_\text{in} - 1) \times S^d_1 + F^d_1 - 2 P^d_1 + O^d_1 \label{eq:const_d_1},\\
    28 = (W^d_\text{out} - 1) \times S^d_2 + F^d_2 - 2 P^d_2 + O^d_2 \label{eq:const_d_2},
\end{align}
i.e. we need to choose the parameters $F^d_1, F^d_2, P^d_1, P^d_2, S^d_1, S^d_2, O^d_1, O^d_2$ such that the output size is 28, which is challenging, as only a small number of parameter configurations fulfill this property.
While this problem is already challenging when assuming a given fixed input image shape $W_\text{in}^d$, in our setting it is more difficult, as $W_\text{in}^d$ has to be of a suitable size as well. Note that $W_\text{in}^d$ is determined by the size of the fully-connected layer preceding the first deconvolutional layer, which yields an additional challenge: the size of the last fully-connected layer has to be set such that it can be resized to an image of shape $C^d_1 \times W_\text{in}^d \times W_\text{in}^d$ (i.e., such that it can be fed into a deconvolutional layer), where $C^d_1$ denotes the number of channels of the first deconvolutional layer of the decoder.
As the resulting problem would be too challenging for any algorithm to produce a valid solution in a reasonable amount of time, we simplify it slightly by only treating $C^d_1$ as a design parameter (as described in \cref{subsec:hypers_vae}), but keeping $W^d_\text{in} = 7$ fixed.
The value 7 is chosen since $16 \times 7 \times 7 = 784$, i.e., when setting $C^d_1 = 16$, the last fully-connected layer has the correct output shape (since $28 \times 28 = 784$ for an \texttt{MNIST} and \texttt{FashionMNIST} image).
This way, a valid decoder architecture can be achieved by deactivating all convolutional layers and choosing $C^d_1 = 16$, constituting an alternative if fulfilling the decoder constraints in Eq.~\eqref{eq:const_d_1} and Eq.~\eqref{eq:const_d_2} is too challenging for an algorithm.

Finally, the constraints in Eq.~\eqref{eq:const_d_1} and Eq.~\eqref{eq:const_d_2} are, respectively, linear and quadratic in the discrete variables $F^d_1, F^d_2, P^d_1, P^d_2, S^d_1, S^d_2, O^d_1, O^d_2$, which again allows us to incorporate them into our optimization routine.

\subsection{Effect of Different Constraint Violation Penalty Values}
\label{subsec:effect_penalty}
We now analyze the effect of the constraint violation penalty value on the performance of SMAC, TPE and GPyOpt. Note that random search and \textsc{MiVaBo} are not affected by the penalty.
We do this analysis to show that the choice of penalty does not qualitatively affect the reported results.
In addition to the penalty of 500 nats considered in the experiments in the main paper, we assessed two smaller alternative penalties of 250 nats and 125 nats, respectively.
The results in \cref{tab:penalties} show that the performance of the methods improves marginally with decreasing penalty values.
This can be intuitively explained by the fact that the smaller the penalty, the smaller the region in hyperparameter space that the penalty discourages from searching.
In fact, a large penalty may not only discourage infeasible configurations, but also feasible configurations that lie "close" to the penalized infeasible one (where closeness is defined by the specific surrogate model employed by the method).
However, even for the smallest penalty of 125 nats, SMAC, TPE and GPyOpt still perform worse than random search, and thus still significantly worse than \textsc{MiVaBo}.
Imposing penalties that are significantly smaller than 125 is not sensible, as this will encourage the model-based methods to violate the constraints, and in turn discourage them from ever evaluating a valid configuration (as this would yield a worse score).

Finally, \cref{tab:violations} shows the number of constraint violations by the different methods, depending on the violation penalty.

\begin{table}[ht!]
    \centering
    \caption{Mean plus/minus one standard deviation of the negative test log-likelihood over 8 random initializations, achieved by the best VAE configuration found by SMAC, TPE and GPyOpt after 16 BO iterations, for constraint violation penalties of 500, 250 and 125 nats. Performance values of \textsc{MiVaBo} and random search (which are not affected by the penalty) are included for reference.}
    \begin{tabular}{ l r r r }
        \toprule
        \textbf{Algorithm} & \multicolumn{3}{c}{\textbf{Penalty (nats)}}\\
        \cline{2-4}
        & \textbf{500} & \textbf{250} & \textbf{125}\\
        \midrule
        \textbf{SMAC} & $113.0 \pm 1.8$ & $112.1 \pm 1.8$ & $111.1 \pm 1.6$\\
        \textbf{TPE} & $108.8 \pm 1.2$ & $108.1 \pm 1.3$ & $108.1 \pm 1.3$\\
        \textbf{GPyOpt} & $108.5 \pm 1.1$ & $108.5 \pm 0.6$ & $106.5 \pm 1.4$\\
        \midrule
        \textbf{RS} & & & $106.3 \pm 0.9$\\
        \textbf{\textsc{MiVaBo}} & & & $\mathbf{94.4 \pm 0.8}$\\
        \bottomrule
    \end{tabular}
    \label{tab:penalties}
\end{table}

\begin{table}[ht!]
    \centering
    \caption{Mean plus/minus one standard deviation of the number of constraint violations by SMAC, TPE, GPyOpt and random search within 16 BO iterations over 8 random initializations, for constraint violation penalties of 500, 250 and 125 nats.}
    \begin{tabular}{ l r r r }
        \toprule
        \textbf{Algorithm} & \multicolumn{3}{c}{\textbf{Penalty (nats)}}\\
        \cline{2-4}
        & \textbf{500} & \textbf{250} & \textbf{125}\\
        \midrule
        \textbf{SMAC} & $37 \pm 21.7$ & $36 \pm 21.9$ & $28 \pm 11.6$\\
        \textbf{TPE} & $67 \pm 21.3$ & $68 \pm 22.2$ & $68 \pm 22.2$\\
        \textbf{GPyOpt} & $36 \pm 19.3$ & $32 \pm 18.0$ & $27 \pm 10.4$\\
        \textbf{Random search} & $71 \pm 25.5$ & $71 \pm 25.5$ & $71 \pm 25.5$\\
        \bottomrule
    \end{tabular}
    \label{tab:violations}
\end{table}

\subsection{Visualization of Reconstruction Quality}
\label{subsec:reconstructions}
\begin{figure}[ht!]
    \centering
    \includegraphics[scale=1.0]{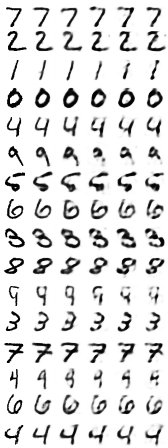}
    \caption{Visualization of the reconstruction quality of a random subset of (non-binarized) images from the \texttt{MNIST} test set, as achieved by the best VAE model (trained for 32 epochs) found by each method. From left to right: ground truth, \textsc{MiVaBo}, random search, GPyOpt, TPE and SMAC. The images are thus ordered (from left to right) by increasing negative test log-likelihood achieved by the VAEs used for reconstruction. Interestingly, the log-likelihood seems to capture quality of visual appearance, as the reconstruction quality may be roughly perceived to decrease from left to right.}
    \label{fig:reconstructions_all}
\end{figure}

While log-likelihood scores allow for a principled quantitative comparison between different algorithms, they are typically hard to interpret for humans.
We thus in \cref{fig:reconstructions_all} visualize the reconstruction quality achieved by the best VAE configuration found by the different methods after 32 BO iterations.
The VAEs were trained for 32 epochs each (as in the BO experiments).
The log-likelihood scores seem to be correlated with quality of visual appearance, and the model found by \textsc{MiVaBo} thus may be perceived to produce the visually most appealing reconstructions among all models.

\section{Discussion and Details on Baselines in Empirical Evaluation}
\label{sec:baselines}
We decided to compare against SMAC \cite{hutter2011} and TPE \cite{bergstra2011a}, as these are state-of-the-art mixed-variable BO methods.
We used their publicly available \texttt{Python} implementations under \url{https://github.com/automl/SMAC3} (SMAC) and \url{https://github.com/hyperopt/hyperopt} (TPE).
We furthermore compare against the popular popular GPyOpt BO \texttt{Python} package \cite{gonzalez2016} (\url{https://github.com/SheffieldML/GPyOpt}) as a reference implementation of a state-of-the-art continuous BO method (which extends to the mixed-variable setting via relaxation and rounding of the discrete variables).
We use these \texttt{Python} packages with their respective default settings.
Moreover, to isolate the benefit of the model choice from the acquisition function optimization procedure, we consider baselines that, respectively, use the \textsc{MiVaBo} and GP models, and optimize the resulting acquisition function using simulated annealing (SA) \cite{kirkpatrick1983}.
For the baseline that combines a GP with simulated annealing, we use the popular GPy \texttt{Python} package \cite{gpy2014} (which also serves as the GP backend of GPyOpt) together with the simulated annealing implementation at \url{https://github.com/perrygeo/simanneal}.
Finally, we compare against random search (using a custom implementation due to its simplicity), which has been shown to be an effective baseline for hyperparameter optimization \cite{bergstra2011a}.

There are several other methods which address problem settings related to the (constrained) mixed-variable paradigm we consider.
We here briefly clarify why we decided to not compare against them in our empirical evaluation.
Firstly, \cite{baptista2018,oh2019,kim2019bayesian} extend BO to tackle purely discrete/combinatorial problems; these approaches can thus not straightforwardly handle the continuous variables present in mixed-variable problems.
\cite{ru2019bayesian} address BO problems with multiple continuous and \emph{categorical} input variables (i.e. unordered ones), whereas \textsc{MiVaBo} includes ordered discrete variables such as integer variables.
As pointed out in \cref{sec:intro}, Hyperband \cite{li2016} and BOHB \cite{falkner2018} are complementary to \textsc{MiVaBo} in that they do not propose new mixed-variable methods, but rather extend existing ones (such as random search and TPE) to the multi-fidelity setting; they should thus not be perceived as competing methods.
The work of \cite{garrido2018} extends GP-based BO to integer variables, but cannot handle discrete constraints.
While several works \cite{hernandez2015c,gardner2014,sui2015safe} propose extensions of continuous BO methods to handle unknown constraints, they can neither handle mixed-variable problems nor known (discrete) constraints, and might thus again be viewed as complementary to our approach.

Finally, a recent line of work extends continuous BO methods to general highly structured input spaces such as graphs or images (which also includes mixed discrete-continuous problems), by first training a deep generative model such as a VAE on the input data, and then using standard continuous BO methods in the continuous latent space learned by the VAE \cite{gomez2018}.
This so-called \emph{latent space optimization} approach has recently been successfully applied to application domains including automatic chemical design and automatic machine learning \cite{gomez2018,kusner2017,nguyen2016synthesizing,luo2018,lu2018,jin2018junction,tripp2020sample}, and might thus be perceived to be a promising method for the mixed-variable hyperparameter tuning tasks we consider in this paper.
However, despite these successes, the latent space optimization paradigm is at an early stage and current methods still suffer from critical shortcomings. 
One of the most severe issues is that the BO procedure tends to progress into regions of the latent space that are too far away from the regions corresponding to the training data, which often results in the BO method suggesting meaningless or even invalid inputs to query (e.g. unreasonable/invalid hyperparameter configurations).
Despite recent efforts attempting to mitigate this issue \cite{kusner2017,griffiths2017,dai2018syntax,daxberger2019bayesian,mahmood2019}, a robust and principled solution has yet to be found.
This issue also reveals that the latent space optimization paradigm makes it difficult to incorporate (discrete) constraints, as the optimization is performed in a learned continuous latent space rather than in the original input space (over which the constraints are defined).
As a result, we decided to not compare against latent space optimization methods at this stage, although we point out that this would be an interesting direction for future work. 

\section{Acquisition Function Optimization with Theoretical Guarantees via Dual Decomposition}
\label{sec:dualdecomposition}
As an alternative to the alternating optimization scheme proposed in \cref{subsec:acquisition}, one can also minimize the acquisition function in \cref{eq:f} via dual decomposition - a powerful approach based on Lagrangian optimization, which has well-studied theoretical properties and has been successfully used for many different problems \cite{komodakis2011,sontag2011,rush2012}. 
Despite its versatility, the core idea is simple: decompose the initial problem into smaller solvable subproblems and then extract a solution by cleverly combining the solutions from these subproblems \cite{komodakis2011}.
This requires the following two components:
(1) A set of \emph{subproblems} which are defined such that their sum corresponds to the optimization objective, and which can each be optimized globally, and (2) a so-called \emph{master} problem that coordinates the subproblems to find a solution to the original problem.
One major advantage of dual decomposition algorithms is that they have well-understood theoretical properties\footnote{For details, we refer the interested reader to \cite{komodakis2011,sontag2011,rush2012}}, in particular through connections to linear programming (LP) relaxations.
In fact, they enjoy the best theoretical guarantees in terms of convergence properties, when compared to other algorithms solving this problem \cite{komodakis2011}.
These theoretical properties further facilitate the convergence analysis of \textsc{MiVaBo} outlined in \cref{subsec:regret_bounds}, making dual decomposition algorithms particularly useful for settings where optimization accuracy is of crucial importance.

We now describe how to devise a dual decomposition for our problem, by demonstrating how it can be reformulated in terms of master- and sub-problems (see \cref{sec:dd_derivation} for a detailed derivation).
For convenience, let us denote the discrete, continuous and mixed parts of Eq.~\eqref{eq:f} by $f^d(\x^d) = {\w^d}^\top \ph^d(\x^d)$, $f^c(\x^c) = {\w^c}^\top \ph^c(\x^c)$ and $f^m(\x^d, \x^c) = {\w^m}^\top \ph^m(\x^d, \x^c)$, respectively, thus resulting in the representation $f(\x) = f^d(\x^d) + f^c(\x^c) + f^m(\x^d, \x^c)$.
First, we note that the discrete $f^d(\x^d)$ and continuous $f^c(\x^c)$ parts of Eq.~\eqref{eq:f} already represent easy to solve subproblems (as we assume to have access to an optimization oracle).
It thus remains to discuss the mixed part $f^m(\x^d, \x^c)$.
As $f^m$ is generally difficult to optimize directly, we assume that it decomposes into a sum $f^m(\x) = \sum_{k=1}^{|F|} f^m_k(\x^d_k, \x^c_k)$ of so-called \emph{factors} $f^m_k: \X^d_k \times \X^c_k \rightarrow \R$, where $\x^d_k \in \X^d_k$ and $\x^c_k \in \X^c_k$ respectively denote subvectors of $\x^d$ and $\x^c$ from the (typically low-dimensional) subspaces $\X^d_k \subseteq \X^d$ and $\X^c_k \subseteq \X^c$.
Here, $F$ denotes a set of subsets $k \in F$ of the variables.
Given this formulation, the initial problem then reduces\footnote{Refer to \cref{sec:dd_derivation} for a detailed derivation.} to the minimization of the dual function $L(\lm)$ w.r.t. Lagrange multipliers $\lm$, i.e., the master problem $\min_{\lm} L(\lm)$, with dual function $L(\lm) = \max_{\x^d} \big\{ f^d(\x^d) + \sum_{k \in F} \lm^d_k \x^d_{|k} \big\} + \max_{\x^c} \big\{ f^c(\x^c) + \sum_{k \in F} \lm^c_k \x^c_{|k} \big\} + \sum_{k \in F} \max_{\x^d_k, \x^c_k} \{ f^m_k(\x^d_k, \x^c_k) - \lm^d_k \x_k^d - \lm^c_k \x_k^c \}$.
Here, the master problem coordinates the $2 + |F|$ maximization subproblems,
where $\x^d_{|k} $and $\x^c_{|k}$ respectively denote the subvectors of $\x^d$ and $\x^c$ containing only the variables of factor $k \in F$, $\lm_k^d$ and $\lm_k^c$ are their corresponding Lagrange multipliers.
Intuitively, by updating the dual variables $\lm$, the master problem ensures agreement on the involved variables between the discrete and continuous subproblems and the mixed factors.
Importantly, the dual function $L(\lm)$ only involves independent maximization over local assignments of $\x^d, \x^c$ and $\x^d_k, \x^c_k$, which are assumed to be tractable.
There are two main classes of algorithms used for the maximization, namely subgradient methods and block coordinate descent \cite{sontag2011}.

\section{Derivation of Dual Decomposition}
\label{sec:dd_derivation}
One interesting interpretation of our acquisition function optimization problem as defined in \cref{subsec:acquisition} is as \emph{maximum a posteriori} (MAP) inference in the undirected graphical model, or Markov random field (MRF) \cite{koller2009}, induced by the dependency graph of the involved variables (i.e. the graph in which vertices correspond to variables, and edges appear between variables that interact in some way).
We take this perspective and devise a dual decomposition to tackle the MAP estimation problem induced by our particular setting (i.e., interpreting our acquisition function as the energy function of the graphical model), following the formulation of \cite{sontag2011}.\footnote{In accordance with the notation in \cite{sontag2011}, we will here denote the factors by $\theta$ instead of $f$ (i.e., in contrast to the main text).}

Consider a graphical model on the vertex set $\mathcal{V} = V_d \cup V_c $, where the vertices $V_d = \{1, \ldots, D_d\}$ and $V_c = \{D_d + 1, \ldots, D_d + D_c\}$ correspond to the discrete and continuous variables $\x^d \in \X^d$ and $\x^c \in \X^c$, respectively.
Furthermore, consider a set $F$ of subsets of both discrete and continuous variables/vertices, i.e., $\forall f \in F: f = (f^d\cup f^c) \subseteq V, \emptyset \neq f^d\subseteq V_d, \emptyset \neq f^c \subseteq V_c$, where each subset corresponds to the domain of one of the factors.

Now assume that we are given the following functions on these factors as well as on all discrete/continuous variables:
\begin{itemize}
    \item A factor $\theta^d(\x^d), \theta^d: \X^d\rightarrow \R$ on all discrete variables
    \item A factor $\theta^c(\x^c), \theta^c: \X^c \rightarrow \R$ on all continuous variables
    \item $|F|$ mixed factors $\theta^m_f(\x^d_f, \x^c_f), \theta^m_f: \X^d_f \times \X^c_f \rightarrow \R$ on subsets $f \in F$ of both discrete and continuous variables, where $\x^d_f \in \X^d_f$ and $\x^c_f \in \X^c_f$ respectively denote subvectors of $\x^d$ and $\x^c$ from the (typically low-dimensional) subspaces $\X^d_f \subseteq \X^d$ and $\X^c_f \subseteq \X^c$, indexed by the vertices contained in $f$
\end{itemize}
The goal of our MAP problem is to find and assignment to all variables $\x^d$ and $\x^c$ which maximizes the sum of the factors:
\begin{equation}
\label{eq:opt_original}
    \text{MAP}(\tht) = \max_{\x} \left\{ \theta^d(\x^d) + \theta^c(\x^c) + \sum_{f \in F} \theta^m_f(\x^d_f,\x^c_f) \right\}
\end{equation}
We now slightly reformulate this problem by duplicating the variables $x^d_i$ and $x^c_j$, once for each mixed factor $\theta^m_f(\x^d_f, \x^c_f)$, and then enforce that these variables are equal to the ones appearing in the factors $\theta^d(\x^d)$ and $\theta^c(\x^c)$, respectively.
Let $x^{df}_i$ and $x^{cf}_j$ respectively denote the copy of $x^d_i$ and $x^c_j$ used by factor $f$.
Moreover, denote by $\x^{df}_f = \{x^{df}_i\}_{i \in f^d}$ and $\x^{cf}_f = \{x^{cf}_j\}_{j \in f^c}$ the set of variables used by factor $f$, and by $\x^F = \{\x^{df}_f, \x^{cf}_f\}_{f \in F}$ the set of all variable copies.
We then get the equivalent (but now constrained) optimization problem
\begin{align}
\label{eq:opt_constrained}
    &\max_{\x, \x^F} \left\{ \theta^d(\x^d) + \theta^c(\x^c) + \sum_{f \in F} \theta^m_f(\x^{df}_f,\x^{cf}_f) \right\}\\
    \text{s.t.} &\quad x_i^{df} = x^d_i, \quad \forall f \in F, i \in f^d\nonumber\\
    &\quad x_j^{cf} = x^c_j, \quad \forall f \in F, j \in f^c \nonumber
\end{align}
To remove the coupling constraints, \cite{sontag2011} now propose to use the technique of \emph{Lagrangian relaxation} and introduce a Lagrange multiplier / dual variable $\lambda_{fi}(x_i)$ for every choice of $f \in F$, $i \in f$ and $x_i$ (i.e. for every factor, for every variable in that factor, and for every value of that variable).
These multipliers may then be interpreted as the \emph{message} that factor $f$ sends to variable $i$ about its state $x_i$.

While this works well if all variables are discrete, in our model we also have continuous variables $x^c_j$, and it is clearly not possible to have a Lagrange multiplier for every possible value of $x^c_j$.
To mitigate this issue, we follow \cite{komodakis2011} and instead only introduce a multiplier $\lambda_{fi}$ for every choice of $f \in F$ and $i \in f$, and model the interaction with the variables as $\lambda_{fi}(x_i) = \lambda_{fi} x_i$ (i.e., the product of a multiplier $\lambda_{fi}$ and variable $x_i$).
Observe that since our goal is to relax the coupling constraints, it is sufficient to introduce one multiplier per constraint.
Since we have a constraint for every factor $f \in F$ and every discrete variable $i \in f^d$ and continuous variable $j \in f^c$ in that factor, our approach is clearly viable.

Note that in contrast to \cite{komodakis2011}, that introduces a set of multipliers for \emph{every} factor / subgraph, we only introduce multipliers for the \emph{mixed} factors $f \in F$. 
This is because in contrast to \cite{komodakis2011}, we do not introduce a full set of variable copies for \emph{every} factor and then couple them to another global set of "original" variables, but we instead only introduce variable copies for the \emph{mixed} factors and couple them to the variables appearing in the \emph{discrete and continuous} factors, which we assume to be the "original" variables instead.
This essentially is the same approach used in \cite{sontag2011}, with the difference that \cite{sontag2011} introduce a singleton factor for each variable (i.e., a factor which depends only on a single variable), which they consider to be the "original" variable.
They then simply couple the variable copies appearing in the \emph{higher-order} factors to the "original" variables appearing in the \emph{singleton} factors.
In contrast, in our formulation we don't introduce singleton factors to model the "original" variables, but instead use the \emph{fully discrete and continuous} factors for this purpose, which clearly works equally well.
Note that as a result of this modeling choice, our optimization problem will be unconstrained, regardless of the number of factors, similar as in \cite{sontag2011}. 
In contrast, \cite{komodakis2011} end up with constraints enforcing that some of the dual variables sum to zero, since they are optimizing out the global set of "original" variables from their objective, while we keep the set of "original" variables within our discrete and continuous factors.
For this reason, we will in contrast to \cite{komodakis2011} later not require a projection step within the subgradient method used to optimize the dual; this is to be detailed further below.

For clarity, we treat discrete and continuous variables distinctly and for factor $f \in F$ denote $\lambda^d_{fi}$ and $\lambda^c_{fj}$ respectively for the Lagrange multipliers corresponding to its discrete variables $i \in f^d$ (or rather, the constraints $x^{df}_i = x^d_i$) and its continuous variables $j \in f^c$ (or rather, the constraints $x^{cf}_j = x^c_j$).
For every factor $f \in F$, we furthermore aggregate its multipliers into the vectors $\lm_f^d = \{ \lambda^d_{fi} \}_{i \in f^d} \in \R^{|\X^d_f|}$ and $\lm_f^c = \{ \lambda^c_{fj} \}_{j \in f^c} \in \R^{|\X^c_f|}$.
The set of all Lagrange multipliers is thus $\lm = \{\lambda^d_{fi}: f \in F, i \in f^d\} \cup \{\lambda^c_{fj}: f \in F, j \in f^c\} = \{\lm^d_f, \lm^c_f\}_{f \in F}$.
We then define the Lagrangian
\begin{align*}
    &L(\lm, \x, \x^F) = \theta^d(\x^d) + \theta^c(\x^c) + \sum_{f \in F} \theta^m_f(\x^{df}_f,\x^{cf}_f)\\
    &+ \sum_{f \in F} \sum_{i \in f^d} \lambda^d_{fi} \left( x^d_i - x_i^{df} \right) + \sum_{f \in F} \sum_{j \in f^c} \lambda^c_{fj} \left( x^c_j - x_j^{cf} \right)\\
    &= \left( \theta^d(\x^d) + \sum_{f \in F} \sum_{i \in f^d} \lambda^d_{fi} x^d_i \right)\\
    &+ \left( \theta^c(\x^c) + \sum_{f \in F} \sum_{j \in f^c} \lambda^c_{fj} x^c_j \right)\\ 
    &+ \sum_{f \in F} \left( \theta^m_f(\x^{df}_f,\x^{cf}_f) - \sum_{i \in f^d} \lambda^d_{fi} x_i^{df} - \sum_{j \in f^c} \lambda^c_{fj} x_j^{cf} \right) \ .
\end{align*}
This results in the following optimization problem:
\begin{align}
\label{eq:opt_lagrange}
    &\max_{\x, \x^F} L(\lm, \x, \x^F)\\
    \text{s.t.} &\quad x_i^{df} = x^d_i, \quad \forall f \in F, i \in f^d\nonumber\\
    &\quad x_j^{cf} = x^c_j, \quad \forall f \in F, j \in f^c \nonumber
\end{align}
Note that the problem in Eq.~\eqref{eq:opt_lagrange} is still equivalent to our (hard) original problem in Eq.~\eqref{eq:opt_original} for any assignment of $\lm$, since the Lagrange multipliers cancel out if all coupling constraints are fulfilled.

To obtain a tractable problem, we thus simply omit the coupling constraints in Eq.~\eqref{eq:opt_lagrange} and define the dual function $L(\lm)$ as
\begin{align*}
    &L(\lm) = \max_{\x, \x^F} L(\lm, \x, \x^F)\\
    &= \max_{\x^d} \left( \theta^d(\x^d) + \sum_{f \in F} \sum_{i \in f^d} \lambda^d_{fi} x^d_i \right)\\
    &+ \max_{\x^c} \left( \theta^c(\x^c)
    + \sum_{f \in F} \sum_{j \in f^c} \lambda^c_{fj} x^c_j \right)\\ 
    &+ \sum_{f \in F} \max_{\x^{df}_f, \x^{cf}_f} \left( \theta^m_f(\x^{df}_f,\x^{cf}_f) - \sum_{i \in f^d} \lambda^d_{fi} x_i^{df} - \sum_{j \in f^c} \lambda^c_{fj} x_j^{cf} \right)
\end{align*}
Note that the maximizations are now fully independent, such that we can (without introducing any ambiguity) simplify the notation for the variables involved in the mixed terms to denote $\x^d_f$ and $\x^c_f$ instead of $\x^{df}_f$ and $\x^{cf}_f$, respectively\footnote{I.e., we replace all variable copies $\x^{df}_f, \x^{cf}_f$ in the mixed terms by the "original" variables $\x^d_f, \x^c_f$.}, resulting in the slightly simpler dual formulation
\begin{align*}
    &L(\lm) = \max_{\x^d} \left( \theta^d(\x^d) + \sum_{f \in F} \sum_{i \in f^d} \lambda^d_{fi} x^d_i \right)\\
    &+ \max_{\x^c} \left( \theta^c(\x^c) + \sum_{f \in F} \sum_{j \in f^c} \lambda^c_{fj} x^c_j \right)\\ 
    &+ \sum_{f \in F} \max_{\x^d_f, \x^c_f} \left( \theta^m_f(\x^d_f,\x^c_f) - \sum_{i \in f^d} \lambda^d_{fi} x_i^d - \sum_{j \in f^c} \lambda^c_{fj} x_j^c \right)
\end{align*}
Let $\x^d_{|f} \in \X^d_f$ and $\x^c_{|f} \in \X^c_f$ respectively denote the subvectors of $\x^d$ and $\x^c$ containing only the variables of factor $f$.
The shorthands (or \emph{reparameterizations} \cite{sontag2011})
\begin{align}
    \bar{\theta}_d^{\lm}(\x^d) &= \theta^d(\x^d) + \sum_{f \in F} \sum_{i \in f^d} \lambda^d_{fi} x^d_i \nonumber\\
    \label{eq:fac_d_short}
    &= \theta^d(\x^d) + \sum_{f \in F} \lm^d_f \x^d_{|f}\\
    \bar{\theta}_c^{\lm}(\x^c) &= \theta^c(\x^c) + \sum_{f \in F} \sum_{j \in f^c} \lambda^c_{fj} x^c_j \nonumber\\
    \label{eq:fac_c_short}
    &= \theta^c(\x^c) + \sum_{f \in F} \lm^c_f \x^c_{|f}\\
    \bar{\theta}_f^{\lm}(\x^d_f, \x^c_f) &= \theta^m_f(\x^d_f,\x^c_f) - \sum_{i \in f^d} \lambda^d_{fi} x_i^d - \sum_{j \in f^c} \lambda^c_{fj} x_j^c \nonumber\\
    \label{eq:fac_m_short}
    &= \theta^m_f(\x^d_f,\x^c_f) - \lm^d_f \x_f^d - \lm^c_f \x_f^c
\end{align}
further simplify the dual function $L(\lm)$ to
\begin{equation}
    L(\lm) = \max_{\x^d} \bar{\theta}_d^{\lm}(\x^d) + \max_{\x^c} \bar{\theta}_c^{\lm}(\x^c) + \sum_{f \in F} \max_{\x^d_f, \x^c_f} \bar{\theta}_f^{\lm}(\x^d_f, \x^c_f) \ .
\end{equation}
First, observe that since we maximize over $\x$ and $\x^F$, the dual function $L(\lm)$ is a function of just the Lagrange multipliers $\lm$.
Note that since $L(\lm)$ maximizes over a larger space (since instead of forcing that there must be one global assignment maximizing the objective, we allow the discrete/continuous potentials to be maximized independently of the mixed potentials, meaning that $\x$ may not coincide with $\x^F$), we have for all $\lm$ that
\begin{equation}
    \text{MAP}(\tht) \leq L(\lm) \ .
\end{equation}
The \emph{dual problem} now is to find the tightest upper bound by optimizing the Lagrange multipliers, i.e.
\begin{equation}
\label{eq:dd_master}
    \min_{\lm} L(\lm)
\end{equation}
We also call the dual problem in Eq.~\eqref{eq:dd_master} the \emph{master problem}, which coordinates the $2 + |F|$ \emph{slave problems} (i.e., one for each factor)
\begin{subequations}
    \begin{alignat}{3}
        \label{eq:dd_slave_d}
        &s^d(\lm) &&= \max_{\x^d} \bar{\theta}_d^{\lm}(\x^d)\\
        \label{eq:dd_slave_c}
        &s^c(\lm) &&= \max_{\x^c} \bar{\theta}_c^{\lm}(\x^c)\\
        \label{eq:dd_slaves_m}
        &s^f(\lm) &&= \max_{\x^d_f, \x^c_f} \bar{\theta}_f^{\lm}(\x^d_f, \x^c_f), \quad \forall f \in F \ .
   \end{alignat}
\end{subequations}
where we refer to $s^d$, $s^c$ and $s^f$ as the discrete slave, the continuous slave, and the mixed slaves, respectively.
Using the notation in Eqs.~\eqref{eq:dd_slave_d}-\eqref{eq:dd_slaves_m}, the dual function further simplifies to
\begin{equation}
    L(\lm) = s^d(\lm) + s^c(\lm) + \sum_{f \in F} s^f(\lm) \ .
\end{equation}
Intuitively, the goal of Eq.~\eqref{eq:dd_master} is as follows: The master problem wants the discrete/continuous slaves to agree with the mixed slaves/factors in which the corresponding discrete/continuous variables appear, and conversely, it wants the mixed slaves to agree with the slaves/factors of the discrete/continuous variables in its scope.
The master problem will thus incentivize the discrete/continuous slaves and the mixed slaves to agree with each other, which is done by updating the dual variables $\lm$ accordingly.

The key property of the function $L(\lm)$ is that it only involves maximization over local assignments of $\x^d, \x^c$ and $\x^d_f, \x^c_f$, which are tasks we assume to be tractable.
The dual thus decouples the original problem, resulting in a problem that can be optimized using local operations.
Algorithms that minimize the approximate objective $L(\lm)$ use local updates where each iteration of the algorithms repeatedly finds a maximizing assignment for the subproblems individually, using these to update the dual variables $\lm$ that glue the subproblems together. 
There are two main classes of algorithms of this kind, one based on a subgradient method and another based on block coordinate descent \cite{sontag2011}.

\putbib
\end{bibunit}


\begin{thebibliography}{}

\bibitem[\protect\citeauthoryear{Abeille \bgroup \em et al.\egroup
  }{2017}]{abeille2017}
Marc Abeille, Alessandro Lazaric, et~al.
\newblock Linear {T}hompson sampling revisited.
\newblock {\em EJS}, 2017.

\bibitem[\protect\citeauthoryear{Baptista and Poloczek}{2018}]{baptista2018}
Ricardo Baptista and Matthias Poloczek.
\newblock Bayesian optimization of combinatorial structures.
\newblock In {\em ICML}, 2018.

\bibitem[\protect\citeauthoryear{Bergstra \bgroup \em et al.\egroup
  }{2011}]{bergstra2011a}
James~S Bergstra, R{\'e}mi Bardenet, Yoshua Bengio, and Bal{\'a}zs K{\'e}gl.
\newblock Algorithms for hyper-parameter optimization.
\newblock In {\em NIPS}, 2011.

\bibitem[\protect\citeauthoryear{Boros and Hammer}{2002}]{boros2002}
Endre Boros and Peter~L Hammer.
\newblock Pseudo-boolean optimization.
\newblock {\em Discrete applied mathematics}, 123(1-3):155--225, 2002.

\bibitem[\protect\citeauthoryear{Burda \bgroup \em et al.\egroup
  }{2016}]{burda2015}
Yuri Burda, Roger Grosse, and Ruslan Salakhutdinov.
\newblock Importance weighted autoencoders.
\newblock In {\em ICLR}, 2016.

\bibitem[\protect\citeauthoryear{Chen and Guestrin}{2016}]{chen2016}
Tianqi Chen and Carlos Guestrin.
\newblock {XGB}oost: A scalable tree boosting system.
\newblock In {\em KDD}, 2016.

\bibitem[\protect\citeauthoryear{Eggensperger \bgroup \em et al.\egroup
  }{2015}]{eggensperger2015}
Katharina Eggensperger, Frank Hutter, Holger~H Hoos, and Kevin Leyton-Brown.
\newblock Efficient benchmarking of hyperparameter optimizers via surrogates.
\newblock In {\em AAAI}, 2015.

\bibitem[\protect\citeauthoryear{Falkner \bgroup \em et al.\egroup
  }{2018}]{falkner2018}
Stefan Falkner, Aaron Klein, and Frank Hutter.
\newblock {BOHB}: Robust and efficient hyperparameter optimization at scale.
\newblock In {\em ICML}, 2018.

\bibitem[\protect\citeauthoryear{Gardner \bgroup \em et al.\egroup
  }{2014}]{gardner2014}
Jacob~R. Gardner, Matt~J. Kusner, Zhixiang Xu, Kilian~Q. Weinberger, and
  John~P. Cunningham.
\newblock Bayesian optimization with inequality constraints.
\newblock In {\em ICML}, 2014.

\bibitem[\protect\citeauthoryear{Garrido{-}Merch{\'{a}}n and
  Hern{\'{a}}ndez{-}Lobato}{2018}]{garrido2018}
Eduardo~C Garrido{-}Merch{\'{a}}n and Daniel Hern{\'{a}}ndez{-}Lobato.
\newblock Dealing with categorical and integer-valued variables in bayesian
  optimization with gaussian processes.
\newblock {\em CoRR}, 2018.

\bibitem[\protect\citeauthoryear{Hastie}{2017}]{hastie2017}
Trevor~J Hastie.
\newblock Generalized additive models.
\newblock In {\em Statistical models in S}. Routledge, 2017.

\bibitem[\protect\citeauthoryear{Hazan \bgroup \em et al.\egroup
  }{2017}]{hazan2017}
Elad Hazan, Adam Klivans, and Yang Yuan.
\newblock Hyperparameter optimization: A spectral approach.
\newblock In {\em ICRL}, 2017.

\bibitem[\protect\citeauthoryear{Hern{\'a}ndez-Lobato \bgroup \em et al.\egroup
  }{2015}]{hernandez2015c}
Jos{\'e}~Miguel Hern{\'a}ndez-Lobato, Michael~A Gelbart, Matthew~W Hoffman,
  Ryan~P Adams, and Zoubin Ghahramani.
\newblock Predictive entropy search for bayesian optimization with unknown
  constraints.
\newblock {\em JMLR}, 2015.

\bibitem[\protect\citeauthoryear{Hutter \bgroup \em et al.\egroup
  }{2011}]{hutter2011}
Frank Hutter, Holger~H Hoos, and Kevin Leyton-Brown.
\newblock Sequential model-based optimization for general algorithm
  configuration.
\newblock In {\em LION}, 2011.

\bibitem[\protect\citeauthoryear{Jenatton \bgroup \em et al.\egroup
  }{2017}]{jenatton2017}
Rodolphe Jenatton, Cedric Archambeau, Javier Gonz{\'a}lez, and Matthias Seeger.
\newblock Bayesian optimization with tree-structured dependencies.
\newblock In {\em ICML}, 2017.

\bibitem[\protect\citeauthoryear{Kingma and Welling}{2014}]{kingma2013}
Diederik~P Kingma and Max Welling.
\newblock Auto-encoding variational bayes.
\newblock In {\em ICLR}, 2014.

\bibitem[\protect\citeauthoryear{Krause \bgroup \em et al.\egroup
  }{2008}]{krause2008}
Andreas Krause, Ajit Singh, and Carlos Guestrin.
\newblock Near-optimal sensor placements in {G}aussian processes: Theory,
  efficient algorithms and empirical studies.
\newblock {\em JMLR}, 2008.

\bibitem[\protect\citeauthoryear{Li \bgroup \em et al.\egroup }{2018}]{li2016}
Lisha Li, Kevin Jamieson, Giulia DeSalvo, Afshin Rostamizadeh, and Ameet
  Talwalkar.
\newblock Hyperband: A novel bandit-based approach to hyperparameter
  optimization.
\newblock {\em JMLR}, 2018.

\bibitem[\protect\citeauthoryear{Mo{\v{c}}kus}{1975}]{movckus1975}
Jonas Mo{\v{c}}kus.
\newblock On {B}ayesian methods for seeking the extremum.
\newblock In {\em Optimization Techniques}, 1975.

\bibitem[\protect\citeauthoryear{Mutn\'y and Krause}{2018}]{mutny2018}
Mojmir Mutn\'y and Andreas Krause.
\newblock Efficient high dimensional bayesian optimization with additivity and
  quadrature fourier features.
\newblock In {\em NIPS}, 2018.

\bibitem[\protect\citeauthoryear{Negoescu \bgroup \em et al.\egroup
  }{2011}]{negoescu2011}
Diana~M Negoescu, Peter~I Frazier, and Warren~B Powell.
\newblock The knowledge-gradient algorithm for sequencing experiments in drug
  discovery.
\newblock {\em INFORMS}, 2011.

\bibitem[\protect\citeauthoryear{Oh \bgroup \em et al.\egroup }{2019}]{oh2019}
Changyong Oh, Jakub~M. Tomczak, Efstratios Gavves, and Max Welling.
\newblock Combinatorial bayesian optimization using graph representations.
\newblock {\em NeurIPS}, 2019.

\bibitem[\protect\citeauthoryear{Rahimi and Recht}{2008}]{rahimi2008}
Ali Rahimi and Benjamin Recht.
\newblock Random features for large-scale kernel machines.
\newblock In {\em NIPS}, 2008.

\bibitem[\protect\citeauthoryear{Rolland \bgroup \em et al.\egroup
  }{2018}]{rolland2018}
Paul Rolland, Jonathan Scarlett, Ilija Bogunovic, and Volkan Cevher.
\newblock High-dimensional {B}ayesian optimization via additive models with
  overlapping groups.
\newblock In {\em AISTATS}, 2018.

\bibitem[\protect\citeauthoryear{Salimans \bgroup \em et al.\egroup
  }{2015}]{salimans2014}
Tim Salimans, Diederik~P Kingma, and Max Welling.
\newblock Markov chain monte carlo and variational inference: Bridging the gap.
\newblock In {\em ICML}, 2015.

\bibitem[\protect\citeauthoryear{Shahriari \bgroup \em et al.\egroup
  }{2016}]{shahriari16}
Bobak Shahriari, Kevin Swersky, Ziyu Wang, Ryan~P. Adams, and Nando de~Freitas.
\newblock Taking the human out of the loop: {A} review of {B}ayesian
  optimization.
\newblock {\em IEEE}, 2016.

\bibitem[\protect\citeauthoryear{Sontag \bgroup \em et al.\egroup
  }{2011}]{sontag2011}
David Sontag, Amir Globerson, and Tommi Jaakkola.
\newblock Introduction to dual composition for inference.
\newblock In {\em Optimization for Machine Learning}. 2011.

\bibitem[\protect\citeauthoryear{Srinivas \bgroup \em et al.\egroup
  }{2010}]{srinivas09}
Niranjan Srinivas, Andreas Krause, Sham~M Kakade, and Matthias Seeger.
\newblock {Gaussian} process optimization in the bandit setting: No regret and
  experimental design.
\newblock In {\em ICML}, 2010.

\bibitem[\protect\citeauthoryear{Sui \bgroup \em et al.\egroup
  }{2015}]{sui2015safe}
Yanan Sui, Alkis Gotovos, Joel Burdick, and Andreas Krause.
\newblock Safe exploration for optimization with gaussian processes.
\newblock In {\em ICML}, 2015.

\bibitem[\protect\citeauthoryear{Thompson}{1933}]{thompson1933}
William~R Thompson.
\newblock On the likelihood that one unknown probability exceeds another in
  view of the evidence of two samples.
\newblock {\em Biometrika}, 1933.

\bibitem[\protect\citeauthoryear{Vanschoren \bgroup \em et al.\egroup
  }{2014}]{vanschoren2014}
Joaquin Vanschoren, Jan~N Van~Rijn, Bernd Bischl, and Luis Torgo.
\newblock Open{ML}: networked science in machine learning.
\newblock {\em ACM SIGKDD}, 15(2):49--60, 2014.

\bibitem[\protect\citeauthoryear{Williams and
  Rasmussen}{2006}]{williams2006gaussian}
Christopher~KI Williams and Carl~Edward Rasmussen.
\newblock {\em Gaussian processes for machine learning}.
\newblock MIT Press Cambridge, MA, 2006.

\end{thebibliography}


\begin{thebibliography}{}

\bibitem[\protect\citeauthoryear{Baptista and Poloczek}{2018}]{baptista2018}
Ricardo Baptista and Matthias Poloczek.
\newblock Bayesian optimization of combinatorial structures.
\newblock In {\em ICML}, 2018.

\bibitem[\protect\citeauthoryear{Bergstra \bgroup \em et al.\egroup
  }{2011}]{bergstra2011a}
James~S Bergstra, R{\'e}mi Bardenet, Yoshua Bengio, and Bal{\'a}zs K{\'e}gl.
\newblock Algorithms for hyper-parameter optimization.
\newblock In {\em NIPS}, 2011.

\bibitem[\protect\citeauthoryear{Bishop}{2006}]{bishop2006}
Christopher~M Bishop.
\newblock {\em Pattern recognition and machine learning}.
\newblock Springer, 2006.

\bibitem[\protect\citeauthoryear{Dai \bgroup \em et al.\egroup
  }{2018}]{dai2018syntax}
Hanjun Dai, Yingtao Tian, Bo~Dai, Steven Skiena, and Le~Song.
\newblock Syntax-directed variational autoencoder for structured data.
\newblock {\em arXiv preprint arXiv:1802.08786}, 2018.

\bibitem[\protect\citeauthoryear{Daxberger and
  Hern{\'a}ndez-Lobato}{2019}]{daxberger2019bayesian}
Erik Daxberger and Jos{\'e}~Miguel Hern{\'a}ndez-Lobato.
\newblock Bayesian variational autoencoders for unsupervised
  out-of-distribution detection.
\newblock {\em arXiv preprint arXiv:1912.05651}, 2019.

\bibitem[\protect\citeauthoryear{Falkner \bgroup \em et al.\egroup
  }{2018}]{falkner2018}
Stefan Falkner, Aaron Klein, and Frank Hutter.
\newblock {BOHB}: Robust and efficient hyperparameter optimization at scale.
\newblock In {\em ICML}, 2018.

\bibitem[\protect\citeauthoryear{Gardner \bgroup \em et al.\egroup
  }{2014}]{gardner2014}
Jacob~R. Gardner, Matt~J. Kusner, Zhixiang Xu, Kilian~Q. Weinberger, and
  John~P. Cunningham.
\newblock Bayesian optimization with inequality constraints.
\newblock In {\em ICML}, 2014.

\bibitem[\protect\citeauthoryear{Garrido{-}Merch{\'{a}}n and
  Hern{\'{a}}ndez{-}Lobato}{2018}]{garrido2018}
Eduardo~C Garrido{-}Merch{\'{a}}n and Daniel Hern{\'{a}}ndez{-}Lobato.
\newblock Dealing with categorical and integer-valued variables in bayesian
  optimization with gaussian processes.
\newblock {\em CoRR}, 2018.

\bibitem[\protect\citeauthoryear{G{\'o}mez-Bombarelli \bgroup \em et al.\egroup
  }{2018}]{gomez2018}
Rafael G{\'o}mez-Bombarelli, Jennifer~N Wei, David Duvenaud, Jos{\'e}~Miguel
  Hern{\'a}ndez-Lobato, Benjam{\'\i}n S{\'a}nchez-Lengeling, Dennis Sheberla,
  Jorge Aguilera-Iparraguirre, Timothy~D Hirzel, Ryan~P Adams, and Al{\'a}n
  Aspuru-Guzik.
\newblock Automatic chemical design using a data-driven continuous
  representation of molecules.
\newblock {\em ACS central science}, 4(2):268--276, 2018.

\bibitem[\protect\citeauthoryear{Gonz{\'a}lez}{2016}]{gonzalez2016}
J~Gonz{\'a}lez.
\newblock {GP}y{O}pt: A {B}ayesian optimization framework in {P}ython, 2016.

\bibitem[\protect\citeauthoryear{{GPy}}{since 2012}]{gpy2014}
{GPy}.
\newblock {GPy}: A gaussian process framework in python.
\newblock \url{http://github.com/SheffieldML/GPy}, since 2012.

\bibitem[\protect\citeauthoryear{Griffiths and
  Hern{\'a}ndez-Lobato}{2017}]{griffiths2017}
Ryan-Rhys Griffiths and Jos{\'e}~Miguel Hern{\'a}ndez-Lobato.
\newblock Constrained bayesian optimization for automatic chemical design.
\newblock {\em arXiv preprint arXiv:1709.05501}, 2017.

\bibitem[\protect\citeauthoryear{Hern{\'a}ndez-Lobato \bgroup \em et al.\egroup
  }{2013}]{hernandez2013generalized}
Daniel Hern{\'a}ndez-Lobato, Jos{\'e}~Miguel Hern{\'a}ndez-Lobato, and Pierre
  Dupont.
\newblock Generalized spike-and-slab priors for bayesian group feature
  selection using expectation propagation.
\newblock {\em The Journal of Machine Learning Research}, 14(1):1891--1945,
  2013.

\bibitem[\protect\citeauthoryear{Hern{\'a}ndez-Lobato \bgroup \em et al.\egroup
  }{2015a}]{hernandez2015c}
Jos{\'e}~Miguel Hern{\'a}ndez-Lobato, Michael~A Gelbart, Matthew~W Hoffman,
  Ryan~P Adams, and Zoubin Ghahramani.
\newblock Predictive entropy search for bayesian optimization with unknown
  constraints.
\newblock {\em JMLR}, 2015.

\bibitem[\protect\citeauthoryear{Hern{\'a}ndez-Lobato \bgroup \em et al.\egroup
  }{2015b}]{hernandez2015expectation}
Jos{\'e}~Miguel Hern{\'a}ndez-Lobato, Daniel Hern{\'a}ndez-Lobato, and Alberto
  Su{\'a}rez.
\newblock Expectation propagation in linear regression models with
  spike-and-slab priors.
\newblock {\em Machine Learning}, 99(3):437--487, 2015.

\bibitem[\protect\citeauthoryear{Hutter \bgroup \em et al.\egroup
  }{2011}]{hutter2011}
Frank Hutter, Holger~H Hoos, and Kevin Leyton-Brown.
\newblock Sequential model-based optimization for general algorithm
  configuration.
\newblock In {\em LION}, 2011.

\bibitem[\protect\citeauthoryear{IBM}{2009}]{cplex2009v12}
IBM.
\newblock User’s manual for {CPLEX}.
\newblock {\em International Business Machines Corporation}, 46(53):157, 2009.

\bibitem[\protect\citeauthoryear{Jin \bgroup \em et al.\egroup
  }{2018}]{jin2018junction}
Wengong Jin, Regina Barzilay, and Tommi Jaakkola.
\newblock Junction tree variational autoencoder for molecular graph generation.
\newblock {\em arXiv preprint arXiv:1802.04364}, 2018.

\bibitem[\protect\citeauthoryear{Kim \bgroup \em et al.\egroup
  }{2019}]{kim2019bayesian}
Jungtaek Kim, Michael McCourt, Tackgeun You, Saehoon Kim, and Seungjin Choi.
\newblock Bayesian optimization over sets.
\newblock {\em arXiv preprint arXiv:1905.09780}, 2019.

\bibitem[\protect\citeauthoryear{Kirkpatrick \bgroup \em et al.\egroup
  }{1983}]{kirkpatrick1983}
Scott Kirkpatrick, C~Daniel Gelatt, and Mario~P Vecchi.
\newblock Optimization by simulated annealing.
\newblock {\em science}, 220(4598):671--680, 1983.

\bibitem[\protect\citeauthoryear{Koller \bgroup \em et al.\egroup
  }{2009}]{koller2009}
Daphne Koller, Nir Friedman, and Francis Bach.
\newblock {\em Probabilistic graphical models: principles and techniques}.
\newblock MIT press, 2009.

\bibitem[\protect\citeauthoryear{Komodakis \bgroup \em et al.\egroup
  }{2011}]{komodakis2011}
Nikos Komodakis, Nikos Paragios, and Georgios Tziritas.
\newblock {MRF} energy minimization and beyond via dual decomposition.
\newblock {\em IEEE transactions on pattern analysis and machine intelligence},
  2011.

\bibitem[\protect\citeauthoryear{Kusner \bgroup \em et al.\egroup
  }{2017}]{kusner2017}
Matt~J Kusner, Brooks Paige, and Jos{\'e}~Miguel Hern{\'a}ndez-Lobato.
\newblock Grammar variational autoencoder.
\newblock In {\em Proceedings of the 34th International Conference on Machine
  Learning-Volume 70}, pages 1945--1954. JMLR. org, 2017.

\bibitem[\protect\citeauthoryear{Li \bgroup \em et al.\egroup }{2018}]{li2016}
Lisha Li, Kevin Jamieson, Giulia DeSalvo, Afshin Rostamizadeh, and Ameet
  Talwalkar.
\newblock Hyperband: A novel bandit-based approach to hyperparameter
  optimization.
\newblock {\em JMLR}, 2018.

\bibitem[\protect\citeauthoryear{Lu \bgroup \em et al.\egroup }{2018}]{lu2018}
Xiaoyu Lu, Javier Gonzalez, Zhenwen Dai, and Neil Lawrence.
\newblock Structured variationally auto-encoded optimization.
\newblock In {\em International Conference on Machine Learning}, pages
  3273--3281, 2018.

\bibitem[\protect\citeauthoryear{Luo \bgroup \em et al.\egroup
  }{2018}]{luo2018}
Renqian Luo, Fei Tian, Tao Qin, Enhong Chen, and Tie-Yan Liu.
\newblock Neural architecture optimization.
\newblock In {\em Advances in Neural Information Processing Systems}, pages
  7827--7838, 2018.

\bibitem[\protect\citeauthoryear{Mahmood and
  Hern{\'a}ndez-Lobato}{2019}]{mahmood2019}
Omar Mahmood and Jos{\'e}~Miguel Hern{\'a}ndez-Lobato.
\newblock A cold approach to generating optimal samples.
\newblock {\em arXiv preprint arXiv:1905.09885}, 2019.

\bibitem[\protect\citeauthoryear{Minka}{2001}]{minka2001}
Thomas~P Minka.
\newblock Expectation propagation for approximate {B}ayesian inference.
\newblock In {\em UAI}, pages 362--369, 2001.

\bibitem[\protect\citeauthoryear{Nguyen \bgroup \em et al.\egroup
  }{2016}]{nguyen2016synthesizing}
Anh Nguyen, Alexey Dosovitskiy, Jason Yosinski, Thomas Brox, and Jeff Clune.
\newblock Synthesizing the preferred inputs for neurons in neural networks via
  deep generator networks.
\newblock In {\em Advances in neural information processing systems}, pages
  3387--3395, 2016.

\bibitem[\protect\citeauthoryear{Nickisch}{2012}]{nickisch2012}
Hannes Nickisch.
\newblock glm-ie: generalised linear models inference \& estimation toolbox.
\newblock {\em Journal of Machine Learning Research}, 13(May):1699--1703, 2012.

\bibitem[\protect\citeauthoryear{Oh \bgroup \em et al.\egroup }{2019}]{oh2019}
Changyong Oh, Jakub~M. Tomczak, Efstratios Gavves, and Max Welling.
\newblock Combinatorial bayesian optimization using graph representations.
\newblock {\em NeurIPS}, 2019.

\bibitem[\protect\citeauthoryear{Optimization}{2014}]{optimization2014}
Gurobi Optimization.
\newblock Gurobi optimizer reference manual.
\newblock {\em http://www.gurobi.com}, 2014.

\bibitem[\protect\citeauthoryear{Ru \bgroup \em et al.\egroup
  }{2019}]{ru2019bayesian}
Binxin Ru, Ahsan~S Alvi, Vu~Nguyen, Michael~A Osborne, and Stephen~J Roberts.
\newblock Bayesian optimisation over multiple continuous and categorical
  inputs.
\newblock {\em arXiv preprint arXiv:1906.08878}, 2019.

\bibitem[\protect\citeauthoryear{Rush and Collins}{2012}]{rush2012}
Alexander~M Rush and MJ~Collins.
\newblock A tutorial on dual decomposition and {L}agrangian relaxation for
  inference in natural language processing.
\newblock {\em Journal of Artificial Intelligence Research}, 45:305--362, 2012.

\bibitem[\protect\citeauthoryear{Seeger and Nickisch}{2008}]{seeger2008b}
Matthias~W Seeger and Hannes Nickisch.
\newblock Compressed sensing and bayesian experimental design.
\newblock In {\em International Conference on Machine Learning (ICML)}. ACM,
  2008.

\bibitem[\protect\citeauthoryear{Seeger and Nickisch}{2011}]{seeger2011}
Matthias~W Seeger and Hannes Nickisch.
\newblock Large scale {B}ayesian inference and experimental design for sparse
  linear models.
\newblock {\em SIAM Journal on Imaging Sciences}, 4(1):166--199, 2011.

\bibitem[\protect\citeauthoryear{Seeger}{2008}]{seeger2008a}
Matthias~W Seeger.
\newblock Bayesian inference and optimal design for the sparse linear model.
\newblock {\em Journal of Machine Learning Research}, 9(Apr):759--813, 2008.

\bibitem[\protect\citeauthoryear{Sontag \bgroup \em et al.\egroup
  }{2011}]{sontag2011}
David Sontag, Amir Globerson, and Tommi Jaakkola.
\newblock Introduction to dual composition for inference.
\newblock In {\em Optimization for Machine Learning}. 2011.

\bibitem[\protect\citeauthoryear{Sui \bgroup \em et al.\egroup
  }{2015}]{sui2015safe}
Yanan Sui, Alkis Gotovos, Joel Burdick, and Andreas Krause.
\newblock Safe exploration for optimization with gaussian processes.
\newblock In {\em ICML}, 2015.

\bibitem[\protect\citeauthoryear{Tripp \bgroup \em et al.\egroup
  }{2020}]{tripp2020sample}
Austin Tripp, Erik Daxberger, and Jos{\'e}~Miguel Hern{\'a}ndez-Lobato.
\newblock Sample-efficient optimization in the latent space of deep generative
  models via weighted retraining.
\newblock {\em arXiv preprint arXiv:2006.09191}, 2020.

\bibitem[\protect\citeauthoryear{Wainwright \bgroup \em et al.\egroup
  }{2008}]{wainwright2008}
Martin~J Wainwright, Michael~I Jordan, et~al.
\newblock Graphical models, exponential families, and variational inference.
\newblock {\em Foundations and Trends{\textregistered} in Machine Learning},
  1(1--2):1--305, 2008.

\end{thebibliography}
\end{document}